\documentclass[runningheads, envcountsame, a4paper]{llncs}

\usepackage{microtype}
\usepackage{graphicx}
\usepackage{subcaption}
\usepackage{booktabs} 
\usepackage{amsmath}
\usepackage{amssymb}
\usepackage{bm}
\usepackage{mwe}
\usepackage[hyphens]{url}
\usepackage{multirow}
\usepackage{microtype}
 \usepackage[misc]{ifsym}

\newlength{\tempdima}
\newcommand{\rowname}[1]
{\rotatebox{90}{\makebox[\tempdima][c]{#1}}}

\newcommand{\rownamehorizontal}[1]
{\makebox[\tempdima][c]{#1}}
\renewcommand{\thesubfigure}{\alph{subfigure}}
\newcommand{\mycaption}[1]
{\refstepcounter{subfigure}\textbf{(\thesubfigure) }{\ignorespaces #1}}

\begin{document}

\title{Effects of Boundary Conditions in Fully Convolutional Networks for Learning Spatio-temporal Dynamics \thanks{Supported by the French "Programme d’Investissements d’avenir" ANR-17-EURE-
0005 and the Natural Sciences and Engineering Research Council of
Canada (NSERC). W.G.P. and M.B. are supportted by the French Direction Générale de l'Armement (DGA) through the AID POLA3 project.}}

\titlerunning{Boundary conditions in spatio-temporal CNNs}
\author{Antonio Alguacil {\Letter} \inst{1,2} \orcidID{0000-0001-6310-8287}  \and
Wagner Gonçalves Pinto\inst{2} \orcidID{0000-0003-2365-4229}  \and
Michael Bauerheim {\Letter} \inst{2,1} \orcidID{0000-0003-2365-4229} \and
Marc C.~Jacob   \inst{3} \orcidID{0000-0001-5946-6487}\and
Stéphane Moreau \inst{1}  \orcidID{0000-0002-9306-8375}
}

\authorrunning{A. Alguacil et al.}

%
\institute{Department of Mechanical Engineering, University of Sherbrooke, 2500, boul. de l'Université, Sherbrooke J1K 2R1 QC, Canada \and
Department of Aerodynamics, Energetics and Propulsion, ISAE-Supaero, 10 Avenue Edouard Belin, 31055 Toulouse, France \\
\email{\{antonio.alguacil-cabrerizo, michael.bauerheim\}[at]isae-supaero.fr}  \and 
Université de Lyon, École Centrale de Lyon, INSA Lyon, Université Claude Bernard Lyon 1, CNRS, LMFA, F-69134, Écully, France}
\toctitle{Effects of Boundary Conditions in Fully Convolutional Networks for Learning Spatio-temporal Dynamics}
\tocauthor{Antonio~Alguacil and Wagner~Gonçalves~Pinto and Michael~Bauerheim and
Marc~C.~Jacob and
Stéphane~Moreau}
\maketitle        
\setcounter{footnote}{0}

\begin{abstract}
Accurate modeling of boundary conditions is crucial in computational physics. The ever increasing use of neural networks as surrogates for physics-related problems calls for an improved understanding of boundary condition treatment, and its influence on the network accuracy. In this paper, several strategies to impose boundary conditions (namely padding, improved spatial context, and explicit encoding of physical boundaries) are investigated in the context of fully convolutional networks applied to recurrent tasks. These strategies are evaluated on two spatio-temporal evolving problems modeled by partial differential equations: the 2D propagation of acoustic waves (hyperbolic PDE) and the heat equation (parabolic PDE). Results reveal a high sensitivity of both accuracy and stability on the boundary implementation in such recurrent tasks. It is then demonstrated that the choice of the optimal padding strategy is directly linked to the data semantics. Furthermore, the inclusion of additional input spatial context or explicit physics-based rules allows a better handling of boundaries in particular for large number of recurrences, resulting in more robust and stable neural networks, while facilitating the design and versatility of such types of 
networks.
\footnote{Datasets, code and supplementary material are available at \url{https://gitlab.isae-supaero.fr/a.alguacil/boundary_conditions_fcn_dyn}}

\end{abstract}

\section{Introduction}
\label{sec:introduction}
Recent advances in deep learning have shown an increased use of neural networks to create surrogate models for physics-related problems. In particular, Convolutional Neural Networks (CNN) have been employed in a wide variety of applications, leveraging their efficient parameter sharing property and their ability to capture long-range spatial correlations. However, most of the existing works limit themselves to one particular problem setup, keeping the same boundary conditions (BCs) throughout the entire training data \cite{Lee2019a}, thus being unable to be generalized to other types of boundary conditions. Ideally, a flexible neural network framework should be able to work with several types of boundary conditions, without the need of retraining the network for each new problem setup. It is thus crucial to understand how boundary conditions are treated by data-driven CNNs in order to improve their generalization capabilities.
The general theme of border effect for CNNs has been broadly studied in the image processing community \cite{Liu2008,Hamey2015}. Still today, such border effects can have a strong influence in state-of-the-art architectures employed in image segmentation \cite{Alsallakh2021}. The usual zero-padding strategy leads to border pixel artifacts and blind spots where the network accuracy drops. Solutions have been proposed to treat borders through separate filters \cite{Innamorati2020} for the edges, corners and inner pixels or to consider the padded pixels as missing information, through the use of Partial Convolution strategies \cite{Liu2018a}. Other works demonstrate how CNNs implicitly learn spatial position \cite{Kayhan,Islam2020}, using the padded pixels to serve as anchors, i.e., as a reference for filter activation in border regions. The use of circular convolution \cite{Schubert2019} eliminates such border effects, but can only be employed on ``panoramic'' datasets.

Yet, the previous works have been devised for image segmentation/classification tasks and it is still unclear whether such studies can be directly transposed for regression and recurrent tasks, as usually encountered when modeling physics. In such contexts, a small error in the boundary prediction may lead to large errors elsewhere in the computational domain. A typical example of such a phenomenon is the development of non-reflecting boundary conditions in computational acoustics \cite{Poinsot1992}, which avoids 
the unphysical reflection of waves back into the computational domain. If left untreated, undesired reflections can pollute the calculated solution. To the author's knowledge, there is a lack of clear results regarding what is the optimal strategy for the treatment of such boundary conditions when employing CNNs for spatio-temporal evolving problems, for which BC errors can propagate and contaminate the whole computational domain.
Indeed, only few works specifically focus on the boundary treatment problem. Some employ explicit rules in relatively simple cases, such as in periodic domains, as in the case of turbulence modeling \cite{Mohan2020}. For more complex boundary treatments, previous works are found in the context of Physical-Informed Neural Networks (PINN) \cite{Raissi2019a} employed in combination with CNN \cite{Gao2021}, where hard constraints are imposed through the padding mechanism. However, only simple Dirichlet or Neumann conditions on static problems are considered, leaving dynamical conditions out of the study.
In the case of spatio-temporal modeling, CNNs and Recurrent Neural Networks (RNN) architectures have been employed indistinctly. Mathieu et al.~\cite{Mathieu2016a} designed a Multi-Scale CNN for video prediction, which was later employed in physics-related applications \cite{Lee2019a,Alguacil2020b}. Fotiadis et al.~\cite{Fotiadis2020} compared recurrent and convolutional approaches in physics-based applications and found that CNNs can be successfully employed as spatio-temporal predictors, with greater accuracy than RNNs and lesser training costs. In all previous cases, the treatment of boundary conditions was not explicitly studied, leaving the effect of BC treatments on such spatio-temporal evolving systems as an open question.

In this paper, the effects of several boundary condition treatments are characterized when modeling a space-time evolving problem using Convolutional Neural Networks. Three strategies are employed for handling the boundary conditions: (i) an implicit treatment through padding only, (ii) adding some extra spatial context to the network input and finally, (iii) explicitly encoding the boundary condition rules into the network output. These strategies are tested on a series of datasets with varying boundary conditions, modeling the two dimensional wave and heat equations.

The main contributions are the following:
\renewcommand{\labelenumi}{(\roman{enumi}) }
\begin{enumerate}
    \item For problems with simple Neumann boundary conditions, 
    padding allows CNNs to efficiently model the physics; yet, 
    mimicking the actual semantics of the dataset makes it only possible for problems with fixed BCs;
    \item The addition of an extra spatial context makes the neural network less sensitive to the choice of padding;
    \item Explicit coding of neural networks gives the best accuracy in the more challenging test cases where dynamical effects are needed to model BCs, allowing more robust predictions by first enforcing physics.
\end{enumerate}

\section{Method}
\label{sec:method}

This section describes the methodology to predict the spatio-temporal dynamics of physics-related quantities using a convolutional neural network. The trained network follows a typical auto-regressive strategy \cite{Geneva2019} to produce time series of high-dimensional state vectors. 
The focus is put on the treatment of boundary conditions in the context of convolutional networks, in order to reproduce the desired physics. Several algorithms for the treatment of such BCs are presented and later evaluated in Section \ref{sec:experiments}.

\subsection{Learning an Auto-Regressive Model}
Dynamical systems can be modeled through a discrete time-invariant model $f$ acting on a delayed state vector ${X^t = \{s^t, s^{t-1}, ..., s^{t-k-1}\} }$ composed of $k$ discrete temporal states $s^i$ which may lie on a high dimensional space. Formally, the discretized time-dynamics read :
\begin{equation}
\label{eq:prob_formulation}
Y^{t+1} = f(X^t), 
\end{equation}
where $Y^{t+1} = \{s^{t+1}\}$ corresponds to the next state in the time series.

In order to generate an approximate model for $f$, a neural network $\hat{f}_{\theta}$, parametrized by its weights and biases $\theta$, is trained on a dataset composed of input-target tuples $\{X^t, Y^{t+1}\}_i$ through a supervised optimisation problem, based on an error metric $\mathcal{L}$, such that
\begin{equation}
    \hat{f}_{\theta} = \underset{f_{\theta}}{\text{arg min}} \sum_i \left\{ \mathcal{L}\left[f_{\theta}(X^t), Y^{t+1}\right]_i \right\}.
\end{equation}

Once an approximate solution is obtained, any time state $s^{T}$ can be reached by employing an auto-regressive iterative strategy on the learned model, namely:
\begin{equation}
    Y^{T} = \underbrace{\hat{f}_{\theta} \circ \hat{f}_{\theta} \circ ... \circ \hat{f}_{\theta} }_{T \hbox{ times}} (X^0)
\end{equation}

where $\circ$ is the function composition operator.

\begin{figure}
    \centering
    \includegraphics[width=0.7\textwidth]{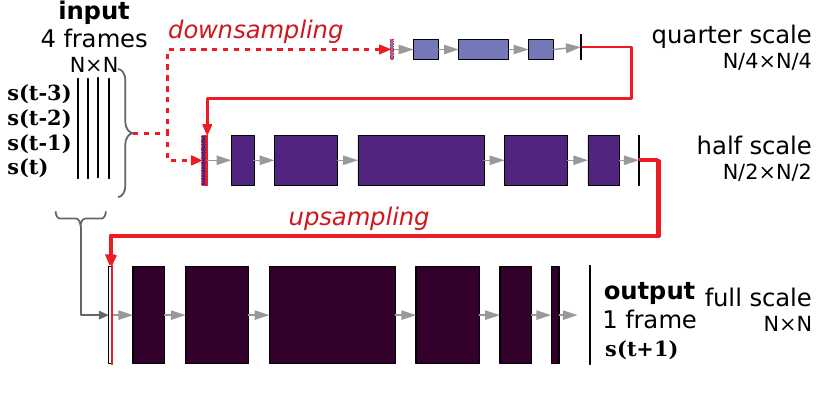}
    \caption{Multi-Scale fully convolutional neural network with 4 consecutive input states of size $N\times N$ and 3 resolution banks ($N/4$, $N/2$ and $N$).  Grey arrows represent 2D convolutions and width of boxes the number of features.}
    \label{fig:multiscale}
\end{figure}

\begin{figure}[t]
\centering
    \begin{subfigure}[b]{0.45\textwidth}
                 \centering
                 \includegraphics[width=\textwidth, ]{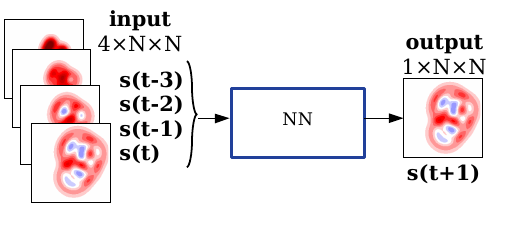}
                 \caption{}
                 \label{fig:methodsA}
         \end{subfigure}
         \begin{subfigure}[b]{0.45\textwidth}
                 \centering
                 \includegraphics[width=\textwidth]{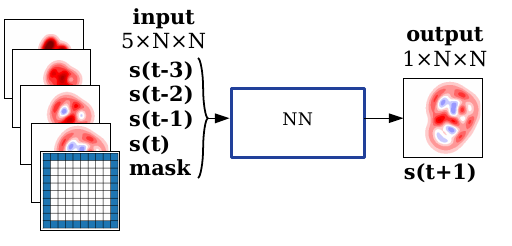}
                 \caption{}
                 \label{fig:methodsB}
         \end{subfigure}%
         
         \begin{subfigure}[b]{0.6\textwidth}
                 \centering
                 \includegraphics[width=\textwidth]{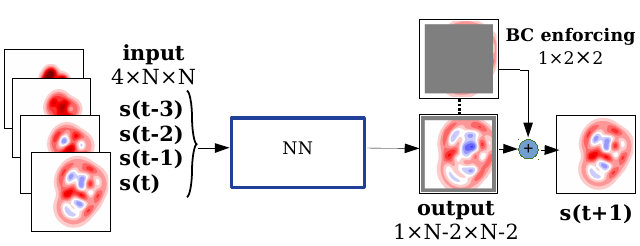}
                 \caption{}
                 \label{fig:methodsC}
         \end{subfigure}%
\caption{Three 
boundary condition strategies:
(a) \textbf{implicit} treatment using only padding,  (b) adding an additional \textbf{spatial context} to the network input or (c) \textbf{explicitly} encoding boundary condition after the neural network prediction.}
\label{fig:methods}
\end{figure}

\subsection{Neural Network Convolutional Architecture}

The auto-regressive strategy can be employed to create surrogate models for physics-based quantities. In traditional fluid solvers, it is common to discretize both the time and space dimensions of physical quantities, such as pressure or velocity fields. This results in high dimensional state where the modeled equations are solved for each degree of freedom.
To reduce the computational costs associated 
with training a neural network surrogate on such high-dimensional states, a convolutional neural network is employed due to its weight-sharing capabilities. In order to efficiently treat the intrinsic multi-scale features of fluid flows, a Multi-Scale fully convolutional neural network \cite{Mathieu2016a,Lee2019a} is employed, as shown in Fig. \ref{fig:multiscale}. 
The input state is composed of four consecutive vectors ${ X^t = \{s^{t-3}, s^{t-2}, s^{t-1}, s^{t}\} }$, in order to provide the network with additional temporal context.

The usage of a pure convolutional approach for the temporal regression problem instead of a Recurent 
Architecture (RNN, LSTMs etc) is justified because of the direct prediction of full states, which only require a short temporal span for their accurate time-stepping (in traditional PDE solvers, the discretization of the time-derivatives). A comparison of the fully convolutional approach with LSTM approaches performed in \cite{Fotiadis2020} confirms this observation, as the convolutional architecture achieves better accuracies on such types of problems.


\subsection{Boundary Condition Treatment}
\label{sec:method_bcs}

In traditional solvers, the 
proper modeling of boundary conditions is key to 
accurate numerical resolution of the partial differential equations. Therefore, understanding the boundary condition treatment is fundamental if convolution neural networks are to be employed as surrogates for physics-based models.
Boundary conditions are intrinsically linked to the concept of padding in fully convolutional networks: additional information must be created at the borders before each convolution in order to keep the same image input resolution at the output. However, the value of the additional pixel information is not known \textit{a priori}, and several padding strategies are available to encode this information. \textbf{Zero padding}, where the additional information is filled with zero values; \textbf{replication padding}, where the values at border pixels are replicated multiple times into the padding area; \textbf{reflection padding}, where an axial symmetry is performed along boundary edges; and \textbf{circular padding} (or \textbf{periodic}) which wraps values from the opposite boundary in the same spatial direction.

While padding is the most straightforward strategy to impose BCs in CNNs, it is unclear how the padding type affects the predictions, neither how an optimal choice is connected to the physical BC type (Dirichlet, Neumann, etc.). Moreover, padding is fixed for the entire database considered, which lacks of versatility when targeting physical systems with multiple possible BCs. Consequently, the objective of this work is to study if an optimal boundary treatment strategy can be found when employing CNNs for physics-based regression. Three types of strategies are considered:

\textbf{Implicit: }It consists in applying exclusively padding to the input and successive feature maps. This is the most common approach in convolutional networks and forces the network to implicitly learn the boundary physics. It is up to the network designer to chose an adequate padding strategy which usually results in a long trial-and-error process until finding an optimal solution. It constitutes the baseline method of this work (Fig. \ref{fig:methodsA}) and the four padding strategies presented above are investigated (\textbf{zeros}, \textbf{replication}, \textbf{reflection} and \textbf{circular}).

\textbf{Spatial context:} The second strategy consists in concatenating an additional channel to the network input, which is consisting in a Boolean mask indicating the position of border pixels. This is formalized as follows:
\begin{equation}
    \mathbb{I}(x)= 
    \begin{cases}
    1, & \mbox{if } x \in \partial \mathcal{D} \\
    0, & \mbox{ otherwise}
    \end{cases}
\end{equation}
where $\mathcal{D}$ represents the domain of interest, $\partial \mathcal{D}$ its boundaries and $x$ are the spatial coordinate. The motivation for such a strategy is to provide the network with an increased spatial context. Figure \ref{fig:methodsB} depicts such a strategy. Note that padding is still employed in order to 
maintain the size of feature maps. This strategy is inspired by other works such as Liu et al.~\cite{Liu2018}, where 
giving explicit spatial information to CNNs 
is shown to crucially improve their generalization capability. Yet, the correlation between the spatial extended context, the padding type, and the actual physical BC is still unclear, thus being studied here.


\textbf{Explicit encoding: } Finally, the third strategy (Fig. \ref{fig:methodsC}) consists in explicitly encoding the boundary condition. In practice, the boundary pixel values are imposed after the network output \cite{Gao2021}, before stepping into the optimization step during the training phase. The way the value is imposed depends on the mathematical modeling of the boundary (Dirichlet, Neumann condition, etc).

All three strategies are compared in this work, by combining them with the four types of padding previously mentioned.

\subsection{Loss Function}
\label{sub:loss}
The loss function for training the aforementioned Multi-Scale network is defined as:
\begin{equation}
\mathcal{L} = \frac{1}{N}\sum_{k=1}^{N} \left\{ \lambda_{L2} \mathcal{L}_2 +  \lambda_{GDL} \mathcal{L}_{GDL} \right\}
\label{eq:loss}
\end{equation}

where $ \mathcal{L}_2 =  || Y^{t+1} - \hat{Y}^{t+1} ||^2_2 $ and $ \mathcal{L}_{GDL} = ||\partial_x Y^{t+1} - \partial_x \hat{Y}^{t+1} ||^2 +  ||\partial_y Y^{t+1} - \partial_y \hat{Y}^{t+1} ||^2 $, where a classical mean square metric is employed, both for the state vector and its spatial derivatives, denoted as Gradient Difference Loss (GDL) as in \cite{Mathieu2016a}. This loss drives the optimization towards achieving sharper predictions, and compensates for the smoothing of the predicted signal in the long term prediction of spatio-temporal series.

Note that the training focuses only on the next time-step prediction. Therefore, the auto-regressive prediction of a long time series of state vectors is a generalization problem.

\section{Applications: Time-evolving PDEs}

The studied modeled is applied to create data-driven surrogates of spatio-temporal evolving partial differential equations. In practice, two applications are investigated:  a hyperbolic PDE (acoustic wave propagation) and parabolic PDE (heat equation). The emphasis is put on the influence of the boundary conditions on the ability of surrogate data-driven models to reproduce accurately the underlying dynamics. Thus, several types of boundary conditions are studied for each case, which are detailed next.


\subsection{Acoustic Propagation of Gaussian Pulses:}
The first application corresponds to the surrogate modelling of a 2D acoustic wave equation in a quiescent medium with speed of sound $c_0$, written in terms of the acoustic density $\rho = \rho(x,y,t)$, with $p$ Gaussian density pulses as initial conditions:

\begin{subequations}
\label{eq:waveEq}
    \begin{equation}
    \frac{\partial^2 \rho}{\partial^2 t} + c_0 \nabla^2 \rho = 0 
    \end{equation}
    \begin{equation}
    \label{eq:waveEqIC}
    \rho(x,y,t=0) = \sum_i^p \varepsilon^i \exp \left\{-\frac{\log 2}{(b^i)^2} \left[ (x-x^i_ 0)^2 + (y-y^i_0)^2 \right] \right\}
    \end{equation}
\end{subequations}
where $\nabla^2$ is the Laplacian operator, $(x^i_0, y^i_0)$, $\varepsilon^i$ and $b^i$ represent respectively the spatial positions of the center, the amplitude and the half-width of the $i^{th}$ initial pulse.

\subsubsection{Boundary Conditions: }
Three cases of boundary conditions are considered, representative of typical configurations found in acoustics. Each one of the BC constitutes a dataset to be employed for training a surrogate model:
\begin{itemize}
    \item {\it Reflecting walls (Dataset 1 - D1)}: Hard-reflecting walls, representing interior acoustics, which is modeled with a Neumann boundary condition: $\bm{\nabla} \rho \cdot \bm{n} = 0$.
    \item{\it Periodic walls (Dataset 2 - D2)}: Periodic conditions to model infinitely repeating domains.
    \item {\it Absorbing walls (Dataset 3 - D3)}: Radiation boundary conditions, modeling propagation of waves into the far-field (exterior acoustics). The challenge 
    is to avoid spurious 
    reflections that can pollute the computational domain.
\end{itemize}

\subsection{Diffusion of Temperature Spots:}
Second, the diffusion of temperature spots is studied, modeled by the following heat equation on the temperature $T = T(x,y,t)$ with $p$ Gaussian density pulses as initial conditions:
\begin{equation}
\label{eq:heatEq}
    \frac{\partial T}{\partial t} + \alpha \nabla^2 T = 0
\end{equation}
where $\alpha$ denotes the thermal diffusivity of the medium. The intial conditions are identical as those employed in Eq. (\ref{eq:waveEqIC}) (Gaussian temperature spots).

\subsubsection{Boundary Conditions: } Here, an additional dataset is generated, called Adiabatic walls (Dataset 4 - D4): Zero-flux adiabatic walls, modeled as Neumann boundary conditions $\bm{\nabla} T \cdot \bm{n} = 0$.

\subsection{Datasets Generation and Parameters}
\label{sec:dataset_param}
The datasets of input-target fields is generated offline with the multi-physics open-source Palabos Lattice-Boltzmann Method (LBM) \cite{Latt2020} numerical solver. We solve equations (\ref{eq:waveEq}) and (\ref{eq:heatEq}) for a duration T with a time-step of $\Delta t$. A two-dimensional square domain is considered.

\paragraph{Acoustic Datasets (D1 to D3)}: Each set is composed of $600$ LBM simulations, each with $T=231$ discrete time snapshots. Only the acoustic density fields $\rho$ are recorded in square domains of physical length $L\times L$, discretized with $N=200$ cells per spatial direction and a spatial step of $\Delta x = L/N = 0.5$. The LB time step is set to $\Delta t_{LBM}= 0.0029 D / c_0 $, with $c_0 = 1/\sqrt{3} \Delta_x / \Delta t_{LBM}$ is the speed of sound. The four input density fields fed into the Neural network are equally spaced in time with $\Delta t_{NN} = \Delta t_{LBM}$. A random number of Gaussian pulses in the range $p \in [1, 5]$ are used as initial conditions, with fixed amplitude and half-width, $\varepsilon = 0.001$ and $b/\Delta x= 12$. The initial location of the pulses $(x_0, y_0)$ is also randomly sampled inside the domain following an uniform distribution. A $500/100$ training/validation random split is employed for the simulations. 

\paragraph{Temperature Datasets (D4)}: Each set is composed of $550$ LBM simulations, each with $T=160$ discrete time snapshots. The temperature fields $T$ are recorded in square domains of physical length $L\times L$, discretized with $N=200$ cells per spatial direction and a spatial step of $\Delta x = L/N = 0.005$. The heat diffusivity is set to $\alpha = 8 \Delta_x^2 / \Delta t_{LBM} $, where the LB time step is set to $\Delta t_{LBM}= 1$. The four input temperature fields fed into the Neural network are sampled so that $\Delta t_{NN} = 4 \Delta t_{LBM}$. The same initial conditions as in D1-D3 are employed. A $400/150$ training/validation random split is employed for the simulations. 

\section{Results}
\label{sec:experiments}

\begin{figure}[htbp!]
\settoheight{\tempdima}{\includegraphics[width=.1\linewidth]{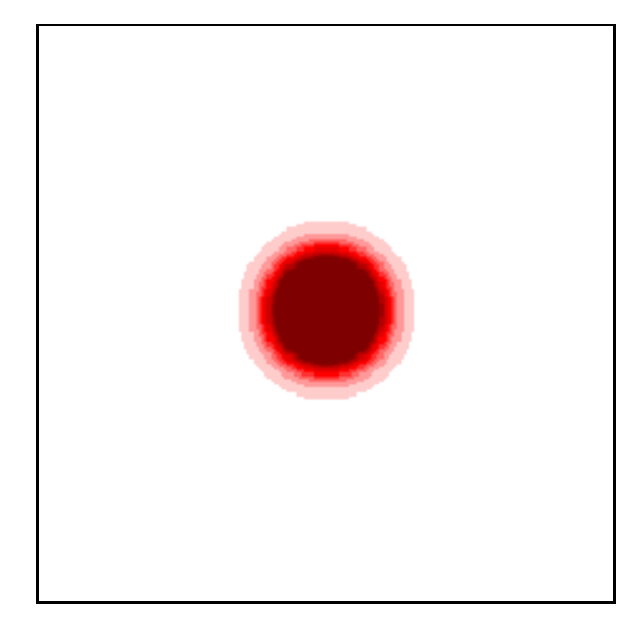}}%
\centering\begin{tabular}{@{}c@{}c@{}c@{}c@{}c@{}c|c@{}c@{ }c@{}c@{}}
& & \multicolumn{4}{c}{Centered IC} & \multicolumn{4}{c}{Random IC} \\
&  &\textbf{it=0} & \textbf{80} & \textbf{160} & \textbf{260} &\textbf{0} & \textbf{40} & \textbf{140} & \textbf{220}\\
\rowname{}&
\rowname{Ref.}&
\includegraphics[width=.1\linewidth]{Figures/d1_refl/impl/center/target/input_0.pdf}&
\includegraphics[width=.1\linewidth]{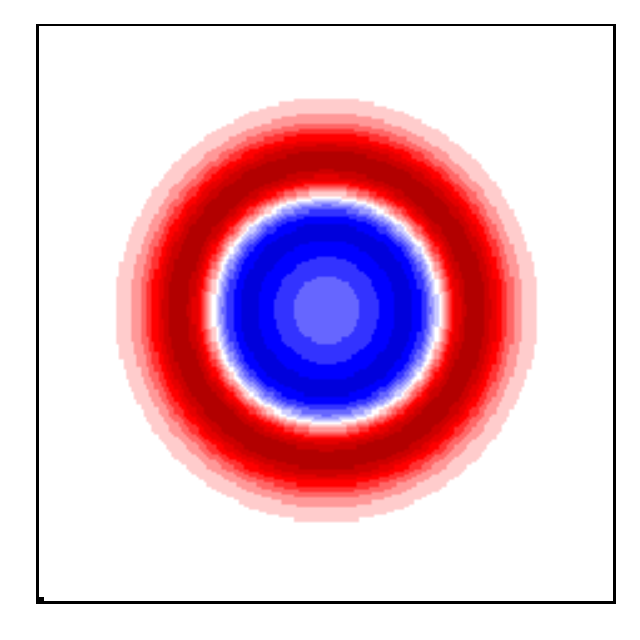}&
\includegraphics[width=.1\linewidth]{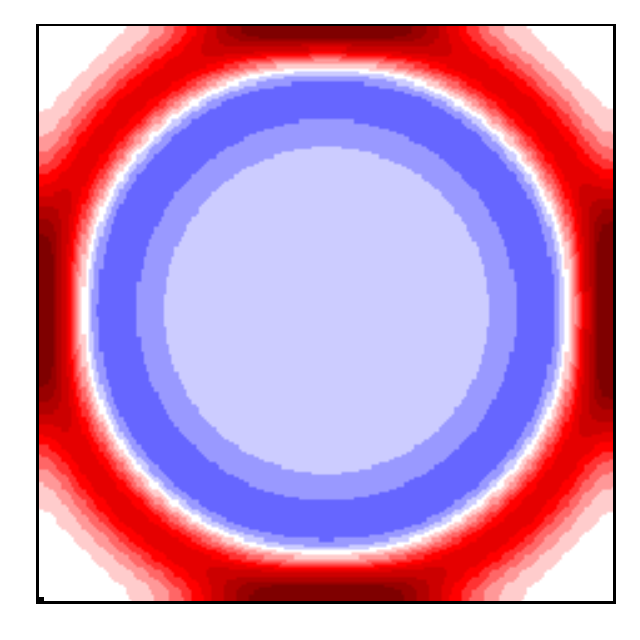}&
\includegraphics[width=.1\linewidth]{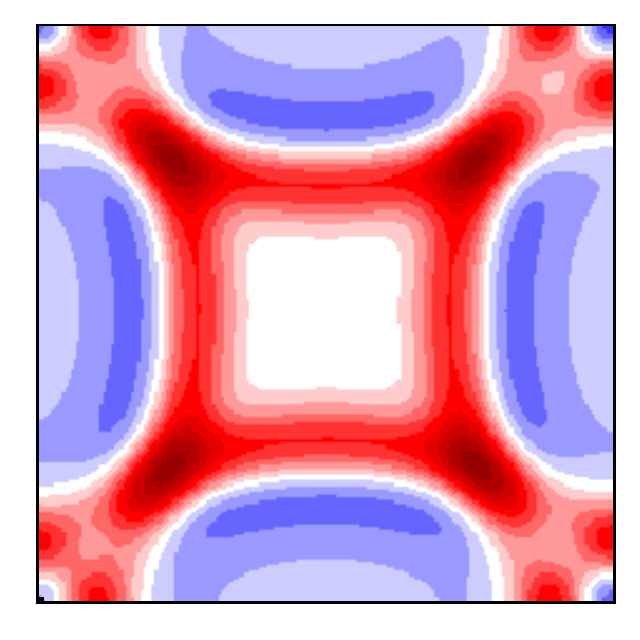}&
\includegraphics[width=.1\linewidth]{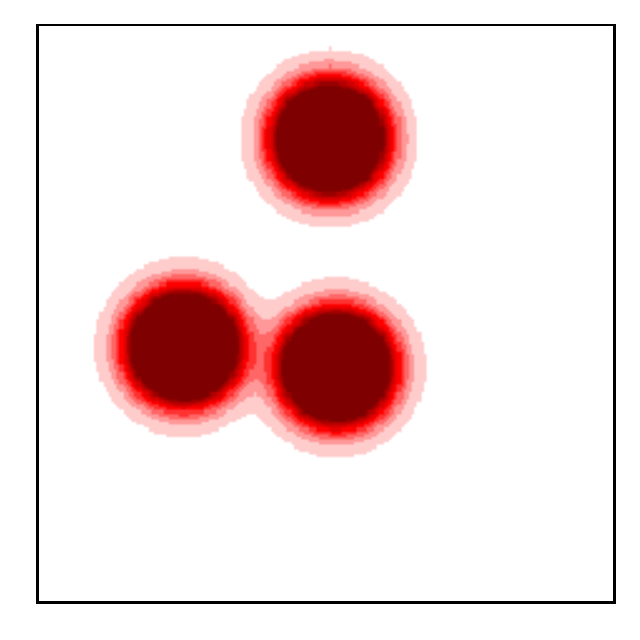}&
\includegraphics[width=.1\linewidth]{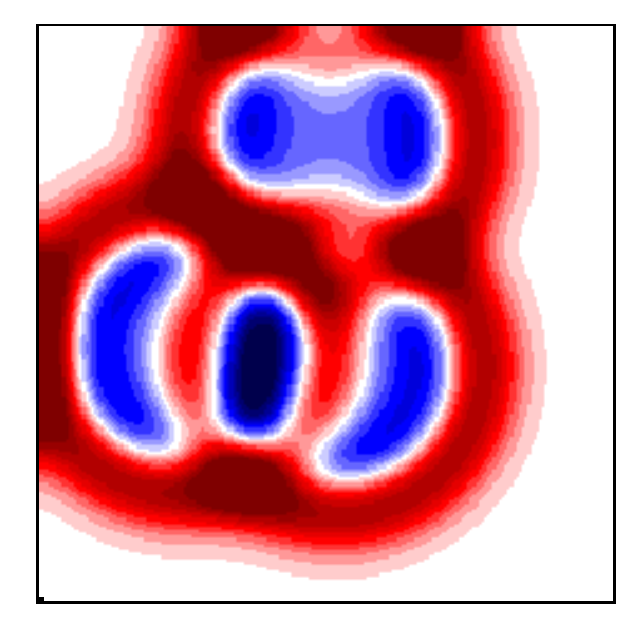}&
\includegraphics[width=.1\linewidth]{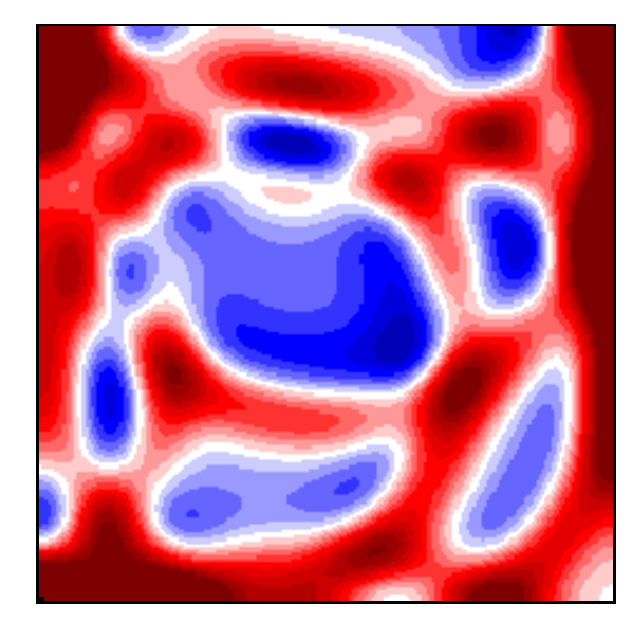}&
\includegraphics[width=.1\linewidth]{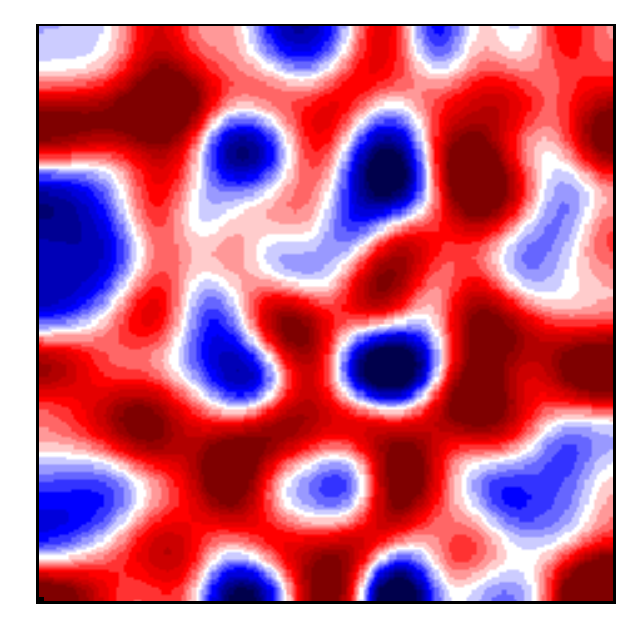}\\[-1ex]
& & & & & \\
\rowname{Impl.}&
\rowname{Repl.}&
\includegraphics[width=.1\linewidth]{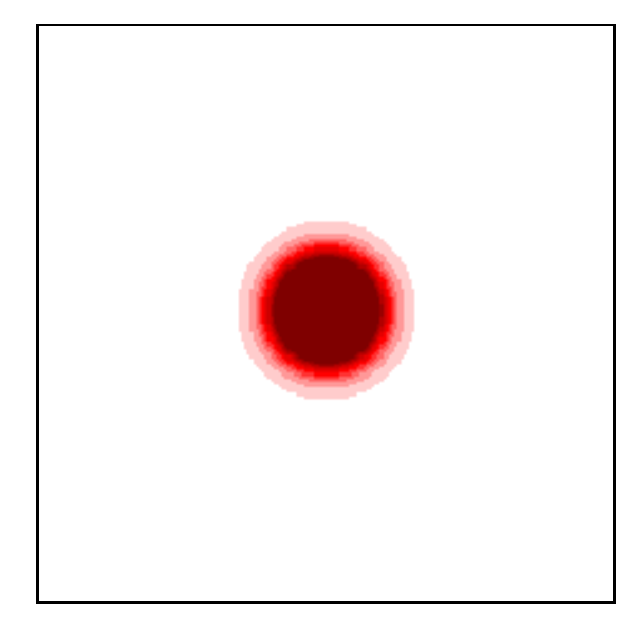}&
\includegraphics[width=.1\linewidth]{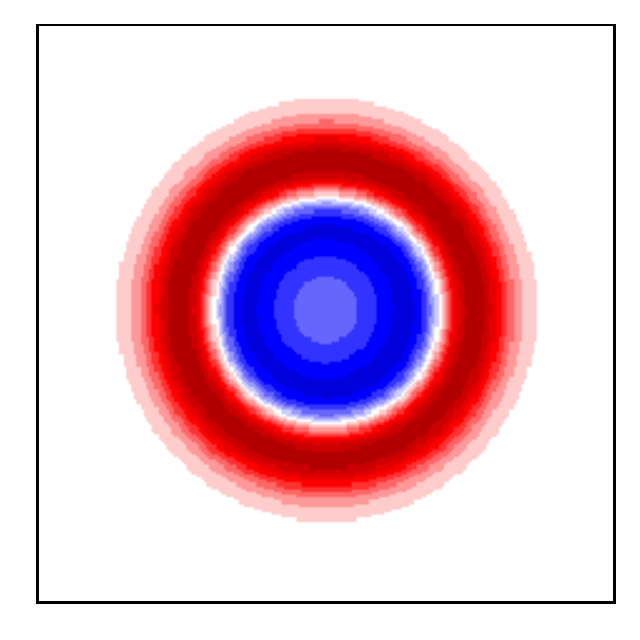}&
\includegraphics[width=.1\linewidth]{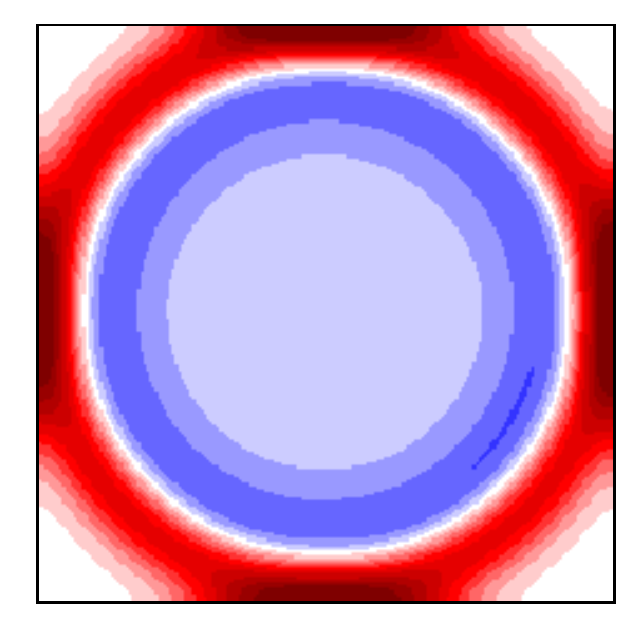}&
\includegraphics[width=.1\linewidth]{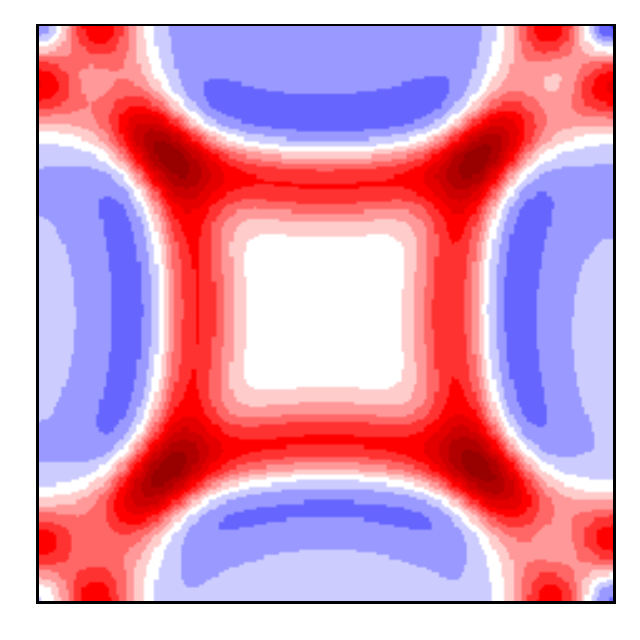}&
\includegraphics[width=.1\linewidth]{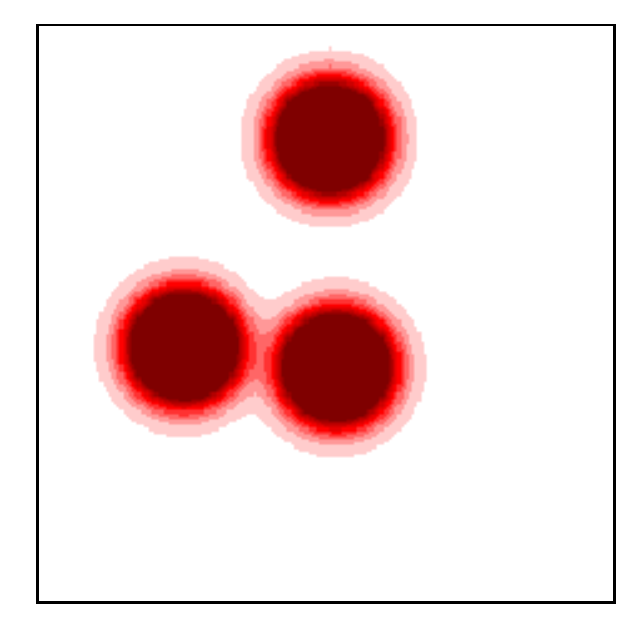}&
\includegraphics[width=.1\linewidth]{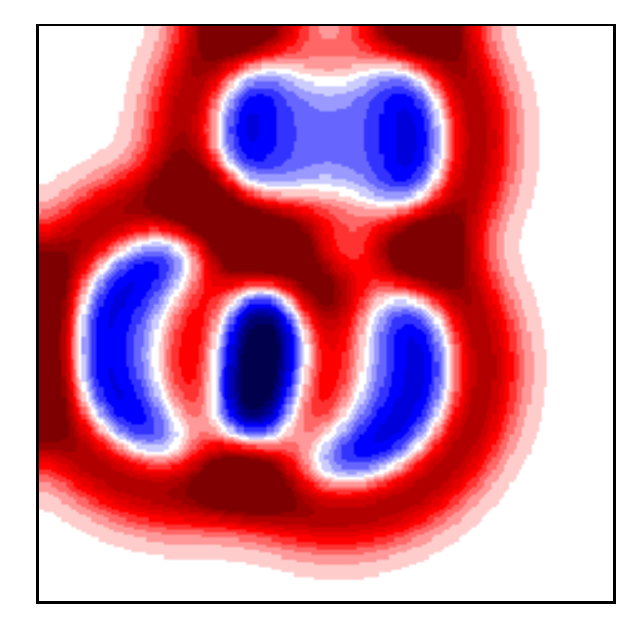}&
\includegraphics[width=.1\linewidth]{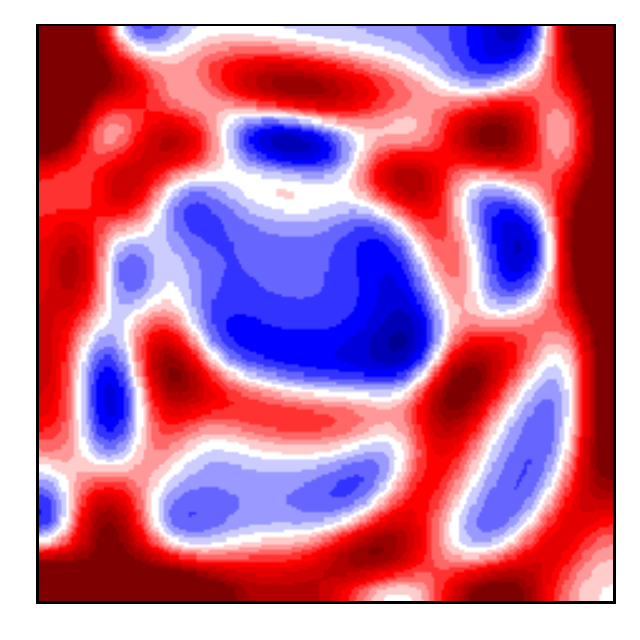}&
\includegraphics[width=.1\linewidth]{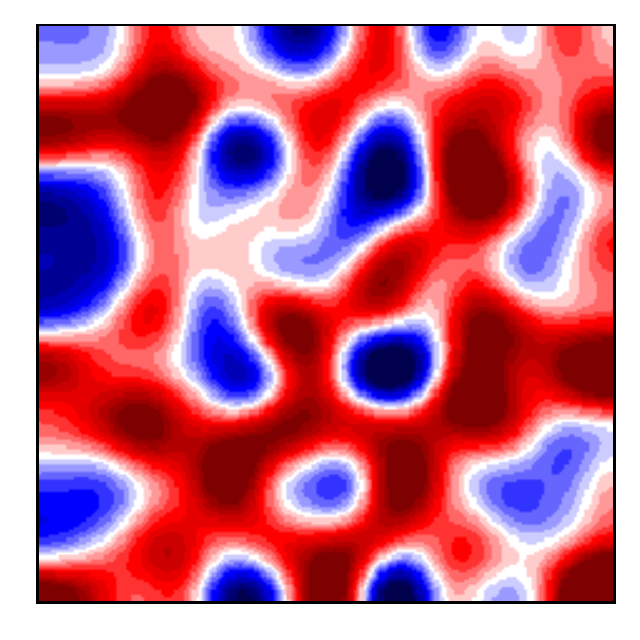}\\[-1ex] & & & & & \\
\rowname{Impl.}&
\rowname{Circ.}&
\includegraphics[width=.1\linewidth]{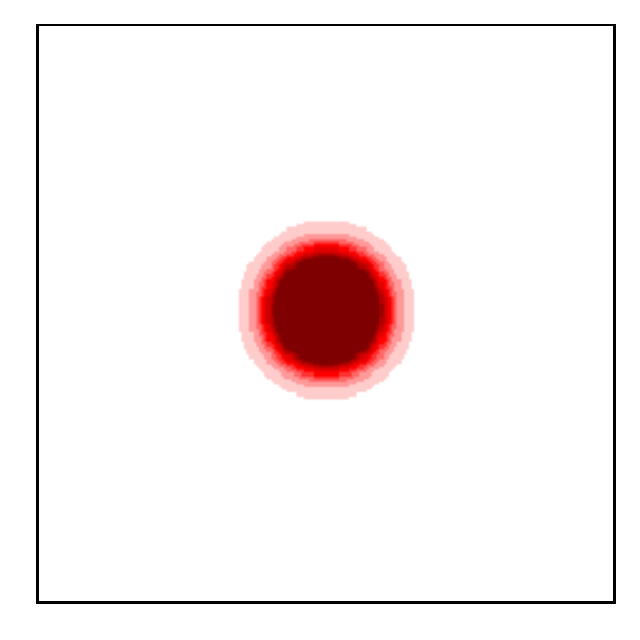}&
\includegraphics[width=.1\linewidth]{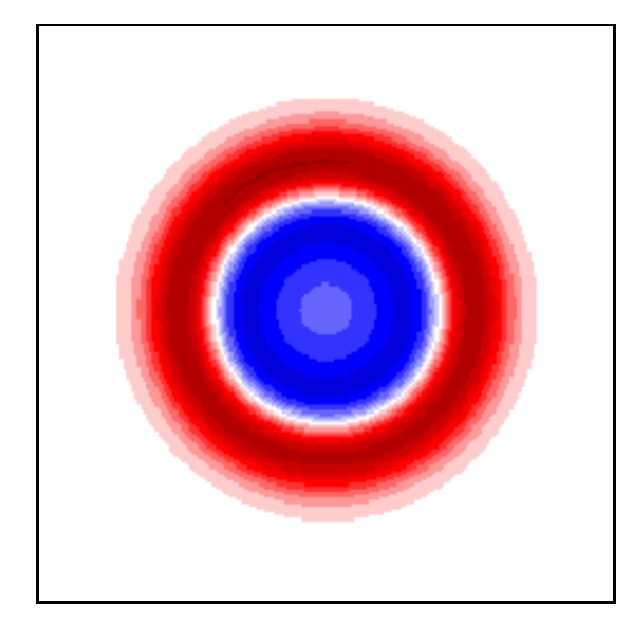}&
\includegraphics[width=.1\linewidth]{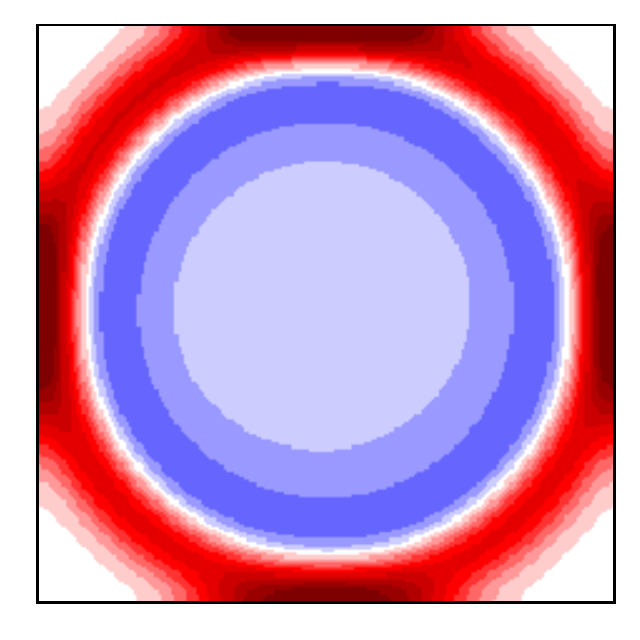}&
\includegraphics[width=.1\linewidth]{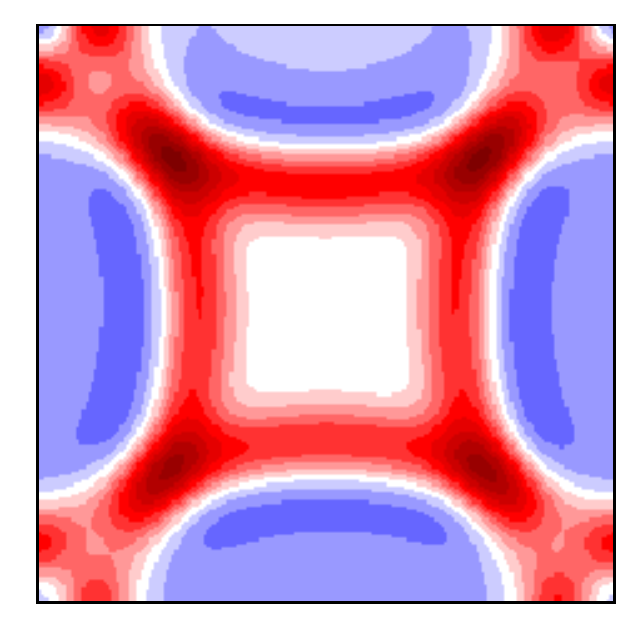}&
\includegraphics[width=.1\linewidth]{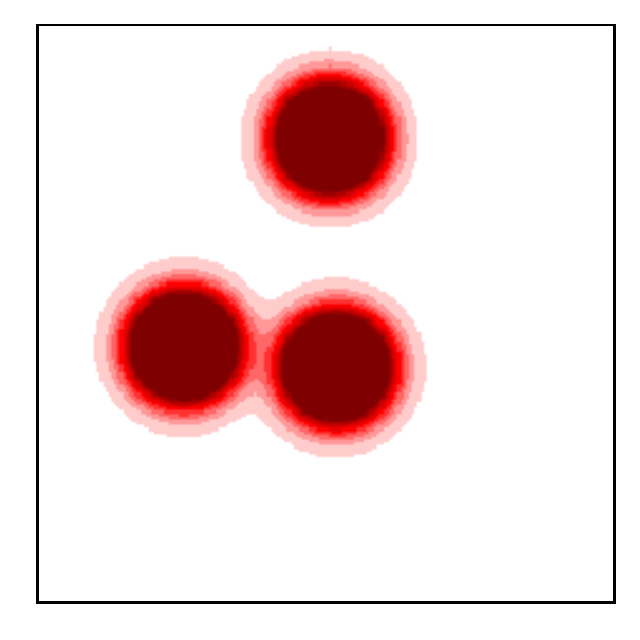}&
\includegraphics[width=.1\linewidth]{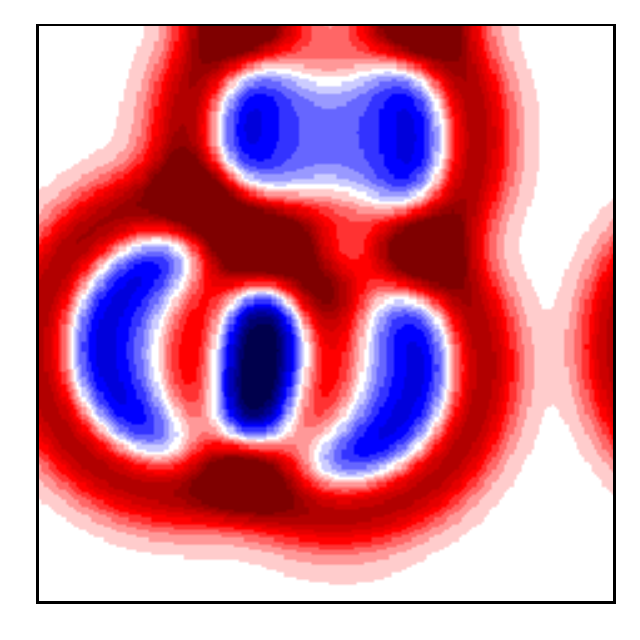}&
\includegraphics[width=.1\linewidth]{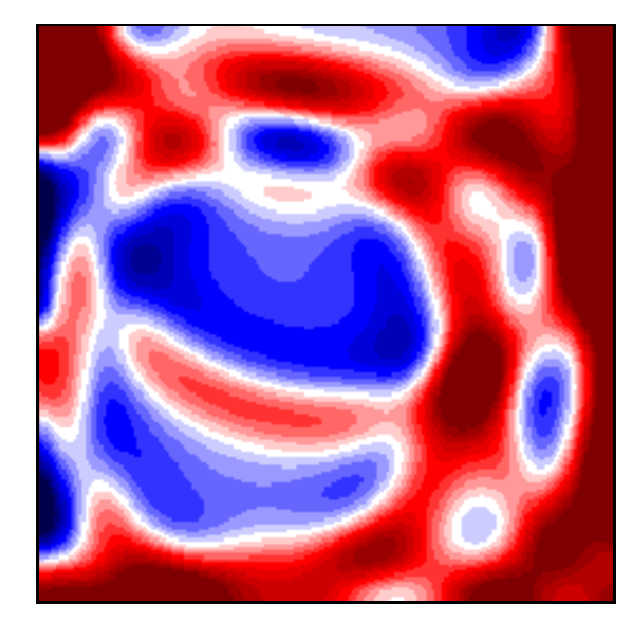}&
\includegraphics[width=.1\linewidth]{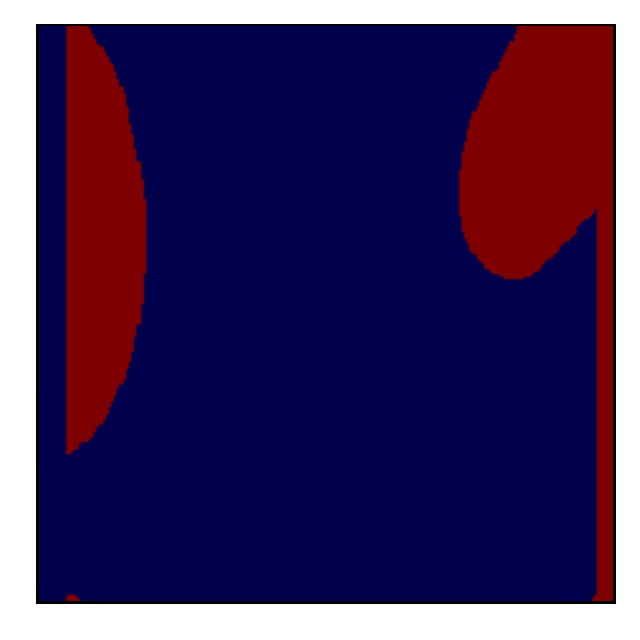}\\[-1ex] & & & & & \\
\rowname{Cont.}&
\rowname{Circ.}&
\includegraphics[width=.1\linewidth]{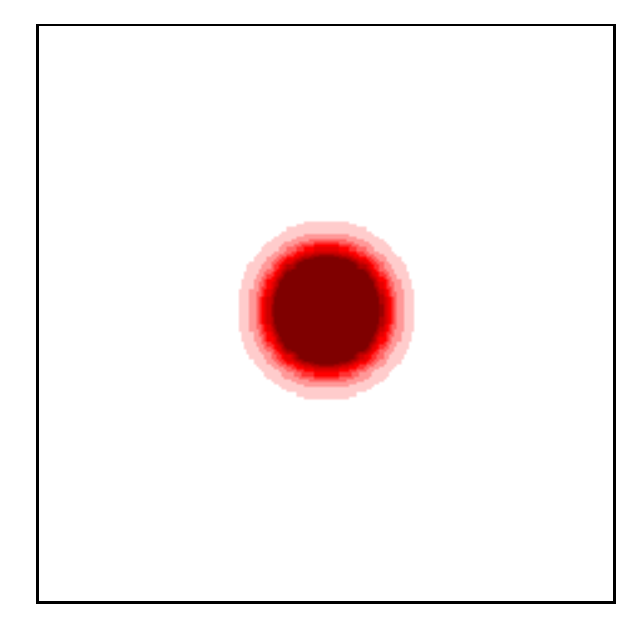}&
\includegraphics[width=.1\linewidth]{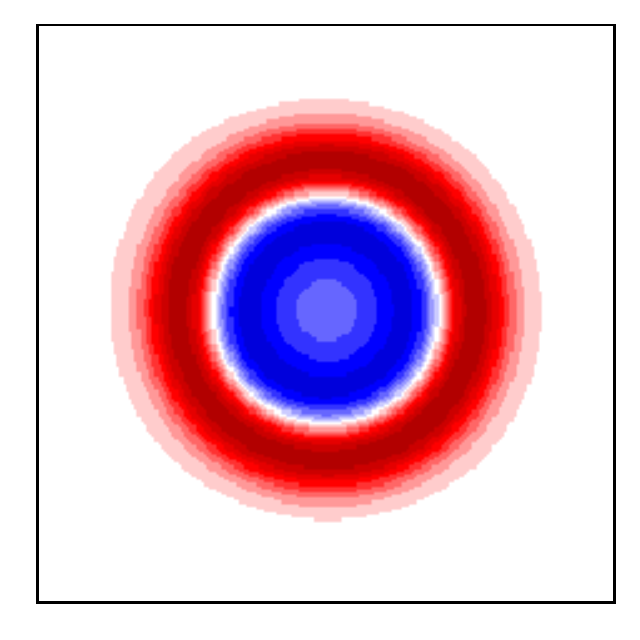}&
\includegraphics[width=.1\linewidth]{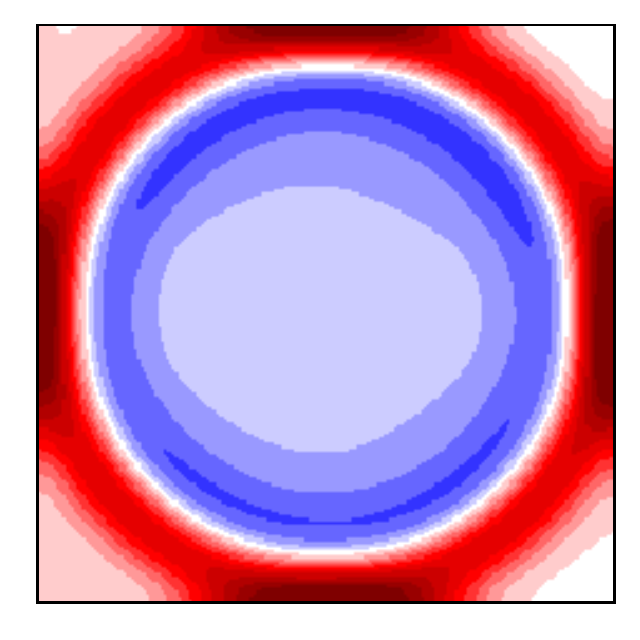}&
\includegraphics[width=.1\linewidth]{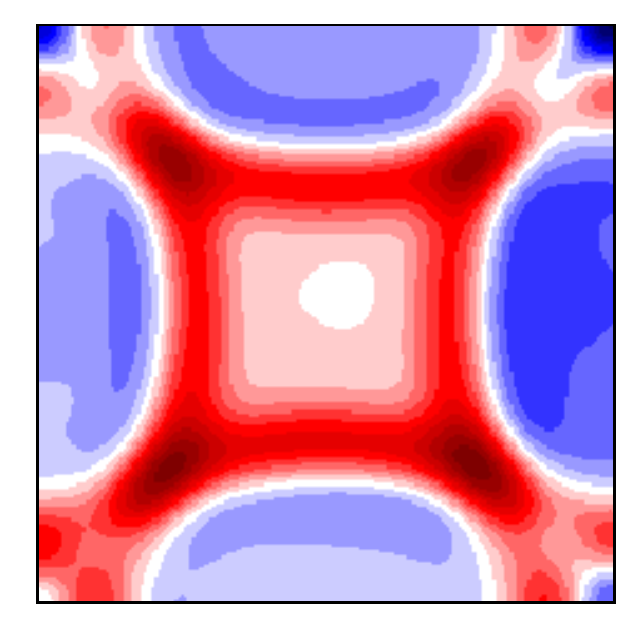}&
\includegraphics[width=.1\linewidth]{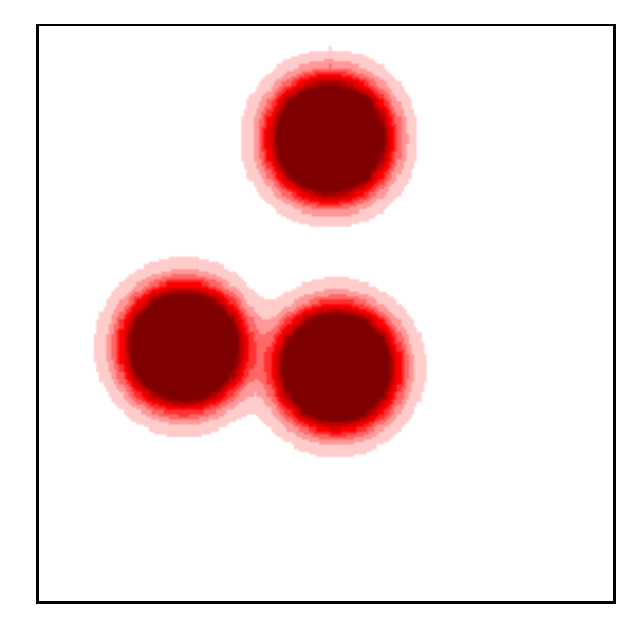}&
\includegraphics[width=.1\linewidth]{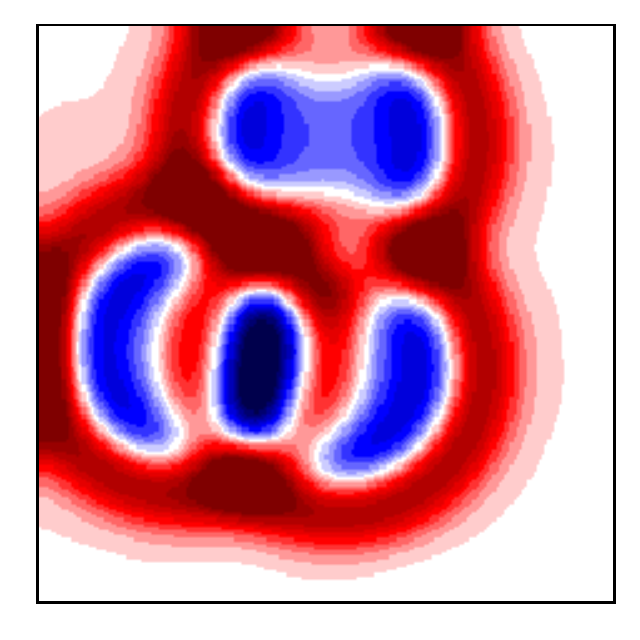}&
\includegraphics[width=.1\linewidth]{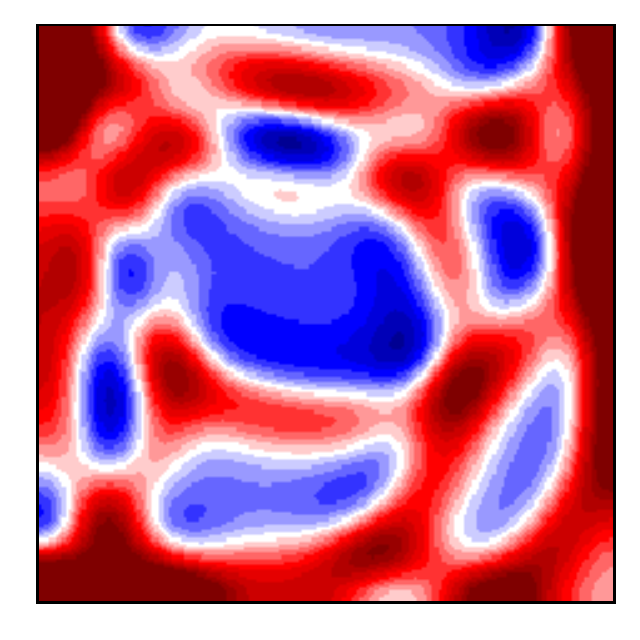}&
\includegraphics[width=.1\linewidth]{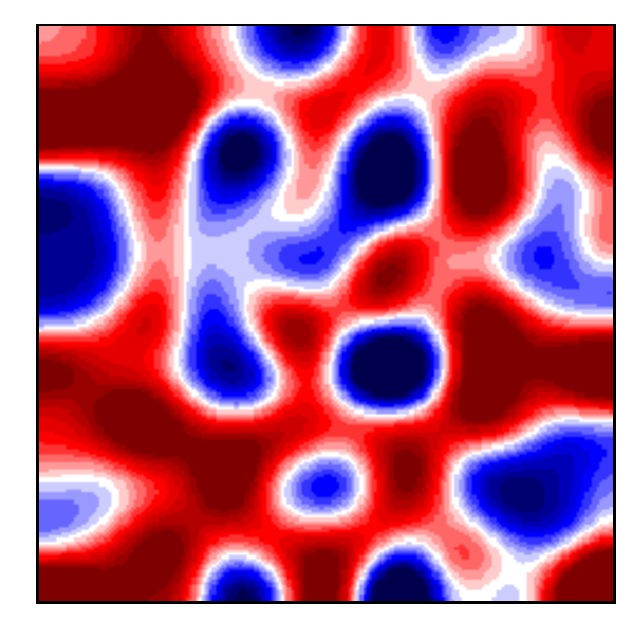}\\
\end{tabular}
\caption{Snapshots of propagating density waves for CNN trained on Dataset 1, for two different initial conditions ($it=0$): centered Gaussian pulse (4 left columns) and randomly sampled Gaussian pulses (4 right columns). Different boundary condition treatments are compared to the LBM reference (top row). For the implicit strategy, the best (resp. worse) results regarding the employed padding are shown in the second row (replication) and third row (circular). For the spatial context strategy, results with circular padding are shown in the last row. }%
\label{fig:results_d1}
\end{figure}

\begin{figure}[hbtp!]
    \centering
    \includegraphics[width=0.9\textwidth]{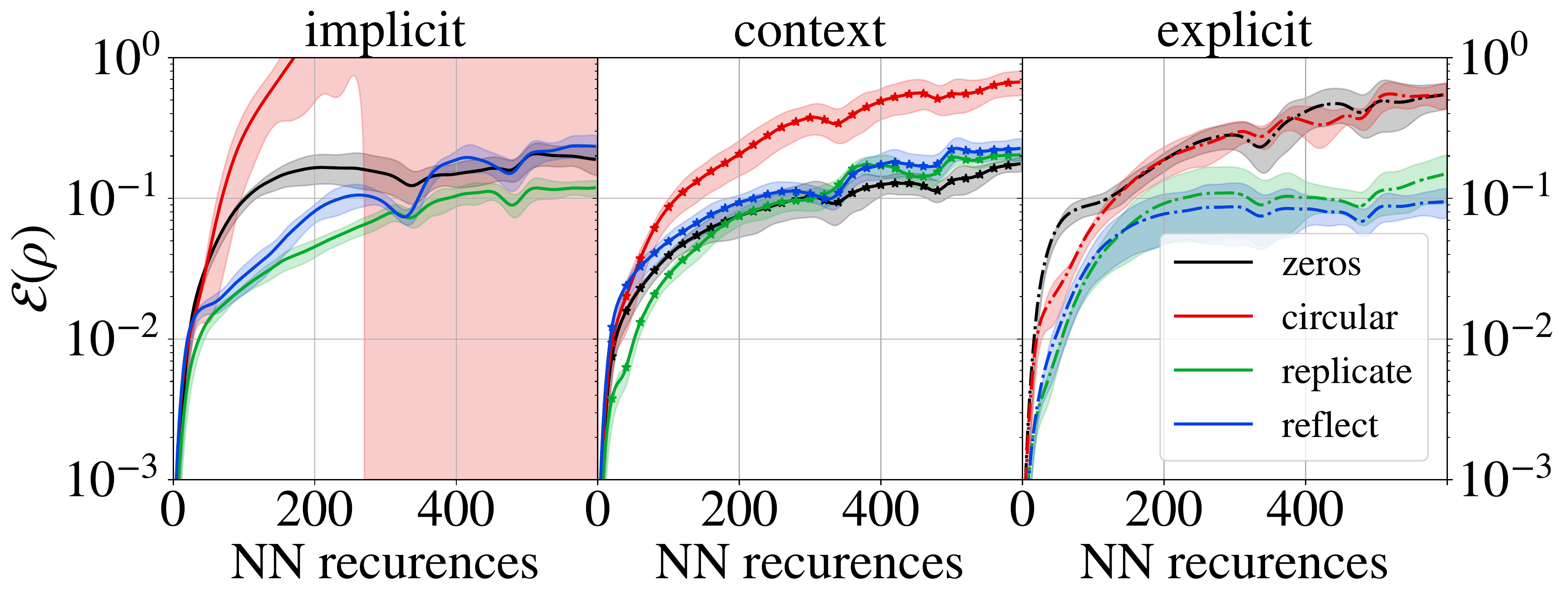}
    \caption{Relative error evolution for case D1 (reflecting walls). 3 boundary treatments are employed: implicit (full lines), spatial context (lines and star marker), and explicit BC treatment (dashed-dotted lines). Results are averaged over 25 different initial conditions and shaded area represent the standard deviation.}
    \label{fig:d1_eror}
\end{figure}

To perform the evaluation of the presented boundary condition treatment, a Multi-Scale network with $0.4$ million parameters is trained on the four proposed datasets. As discussed in \ref{sub:loss} the network is trained to  minimize Eq. \ref{eq:loss} for a single-step prediction. The Adam optimizer is employed, with a learning rate initially set to $10^{-4}$, and decaying by $20\%$ each time the loss reaches a plateau. The loss weights are set to $\lambda_{L2}=0.02$ and $\lambda_{GDL}=0.98$. Data augmentation is employed through the random rotation of input-target tuples, and input fields are normalized by their mean and standard deviation. Batch size is kept at 32. Trainings are performed on a NVIDIA V100 GPU and convergence is achieved at about 1000 epochs for each run. 

\subsubsection{Wave Equation with Reflecting Boundary}: The first case of study corresponds to Dataset 1. Density waves are fully reflected back into the domain after the interaction with boundaries. Therefore, the waves stay in the computational domain for infinitely long times, as no viscous dissipation is present. To test the presented approaches, 24 initial conditions with 1 to 5 Gaussian pulses randomly located in the initial domain are used as inputs for the auto-regressive model. A 25th initial condition is also tested, with the particular case of a Gaussian pulse initially centered in the domain, whose solution leads to strong symmetric solutions and is thus challenging for the neural network. For each initial condition (generated with the LBM solver), the auto-regressive strategy can recursively predict the density fields over a time horizon of $T$ iterations, by using the previous prediction as a new input. In order to improve the neural network robustness versus the error accumulation over time, an \textit{a posteriori} correction is employed to improve the predictions after each time step \cite{Alguacil2020b}, based on the conservation of acoustic energy over time in this particular application.

\begin{table*}[]
  \caption{Averaged relative error for Dataset 1 (hard reflecting walls) at iteration $it=600$ for 25 random initial conditions. Bolded results represent best padding for each strategy.  }
    \begin{small}
      \begin{sc}
        \begin{tabular}{ccccccccccccc}
          \toprule
          & \multicolumn{12}{c}{Padding} \\
          \cmidrule{2-13}
          \multirow{2}{*}{Method} &
          \multicolumn{3}{c}{Zeros} & \multicolumn{3}{c}{Circular} & \multicolumn{3}{c}{Replicate} & \multicolumn{3}{c}{Reflect} \\
          &
          min & max & avg &
          min & max & avg &
          min & max & avg &
          min & max & avg
          \\
          \midrule
          \multicolumn{1}{c|}{Implicit} &
          0.118 & 0.271 & \multicolumn{1}{c|}{0.189} &
          0.306 & $\infty$ & \multicolumn{1}{c|}{$\infty$} &
          0.098 & 0.145 & \multicolumn{1}{c|}{\textbf{0.119}} &
          0.133 & 0.319 & 0.234
          \\
          \multicolumn{1}{c|}{Context} &
          0.114 & 0.213 & \multicolumn{1}{c|}{\textbf{0.176}} &
          0.134 & 0.453 & \multicolumn{1}{c|}{0.670} &
          0.161 & 0.235 & \multicolumn{1}{c|}{0.204} &
          0.138 & 0.297 & 0.226
          \\
          \multicolumn{1}{c|}{Explicit} &
          0.768 & 0.580 & \multicolumn{1}{c|}{1.392} &
          0.291 & 0.780 & \multicolumn{1}{c|}{0.545} &
          0.079 & 0.341 & \multicolumn{1}{c|}{0.148} &
          0.064 & 0.157 & \textbf{0.094}
          \\
          \bottomrule
        \end{tabular}
      \end{sc}
    \end{small}
  \label{tab:d1_results}
\end{table*}

The error is evaluated in terms of relative root mean square error at each neural network iteration, namely $\mathcal{E}(\rho) = \sqrt{||\rho_{t} - \hat{\rho}_{t}||_2} / \sqrt{|| \rho_{t}  ||_2}$, where  $\hat{\rho}_{t}$ is the high-dimensional density prediction at iteration $t$. Here, the time horizon is set at $T=600$ iterations. For the explicit method, Neumann boundary conditions on the density fields are used to model such conditions. A first-order finite difference discretization is employed.

Results are qualitatively evaluated in Fig. \ref{fig:results_d1}. For two different initial conditions (centered pulse and 3 randomly sampled pulses), the LBM reference (top row) is compared to several of the proposed approaches. For the implicit case (i.e., only padding) strategies, the best model regarding the employed padding is shown in the second row, corresponding to the replication padding. For both initial conditions, the auto-regressive strategy follows closely the ground truth data. In the third row, the implicit strategy employs circular padding and shows a good agreement in the case of the initially centered Gaussian pulse. However, for the other initial condition, the prediction diverges after some iterations. At iteration $it=80$, the pulse arriving on the left wall is re-injected at the right wall, mimicking the behavior of periodic boundaries instead of the reflecting ones, on which the network has been trained.


Figure \ref{fig:d1_eror} shows the time-evolution of the error averaged over 25 initial conditions for the different evaluated methods and Table \ref{tab:d1_results} presents the error values for the last prediction at $it=600$. For the implicit strategy, results show that choosing the optimal padding crucially depends on the data physics. With circular padding the network is incapable of reproducing the desired physics except for some particular initial conditions, illustrated by the increased variance area signaling the presence of outliers. This is due to the artifacts discussed previously. For the rest of available padding (zeros, replication and reflection), the replication padding solution performs better than the other two even if error levels remain acceptable (below $2\%$ relative error).

Such observations agree with other studies performed in image segmentation \cite{Alsallakh2021}: circular padding limits the CNNs ability to encode position information and can only be used with spatially periodic data. To further investigate this claim, an additional spatial context channel is employed, while maintaining circular padding. As observed in Fig. \ref{fig:d1_eror} (middle plot, red curve), the error is significantly lower than the one obtained with the implicit strategy. This suggests that the additional input serves as an spatial anchor for the CNN to encode the hard-reflecting wall, which was not possible using only circular padding. The last row in Fig. \ref{fig:results_d1} demonstrates this improvement, even though the prediction error is higher than the one obtained with other padding methods. Furthermore, the addition of the spatial context channel reduces the overall variability of the chosen padding effects. While the error slightly increases for the replication case versus the implicit strategy, all three padding methods converge to similar error evolutions. This suggests that the additional context channel may force the network to explicitly learn similar convolutional kernels for boundary treatment, while this is not guaranteed by the implicit case.

For the explicit case, results in Fig. \ref{fig:d1_eror} (right) show that the replicate and reflection padding cases achieve the lowest errors, while zero and circular paddings have larger errors. 
Note that the network is only trained for a one-step prediction and the explicit enforcing of the boundary is performed after each prediction. Thus, the explicit enforcing is only processed by the network in the auto-regressive context. Results show whether the employed padding strategy is compatible with the enforced boundary values: reflecting and replication padding can be though as first-order finite difference approximations of spatial derivatives, for 1-pixel padding. Zero and circular paddings are on the other hand not compatible with the enforced boundary values, performing worse in both cases.


This behavior highlights the possible benefits of employing explicitly boundary rules as long as the subsequent padding follows the same logic. It also calls to directly enforce such explicit rules in the input padding mechanism, which is left for future work.
\subsubsection{Wave Equation with Periodic BCs}: The second case of study corresponds to Dataset 2, where all four wall boundaries are set as periodic walls in the training data. The objective is that the neural network reproduces the wave propagation in an infinitely-repeating domain.

\begin{figure}[t]
\begin{subfigure}[]{0.31\textwidth}
    \centering
    \includegraphics[width=\textwidth]{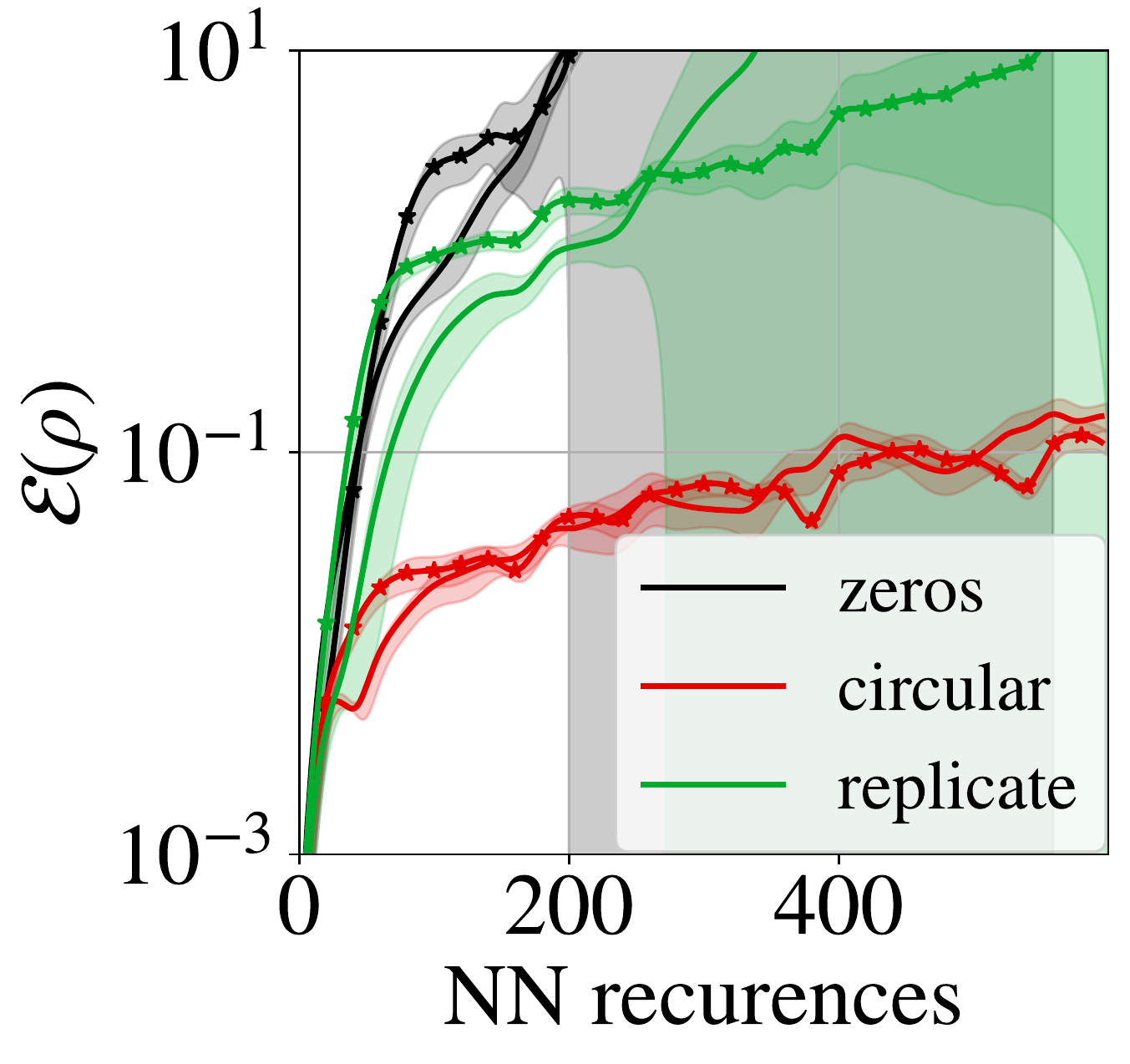}
    \caption{}
\label{fig:results_d2A}
\end{subfigure}
\begin{subfigure}[]{0.68\textwidth}
    \centering
    \includegraphics[width=\textwidth]{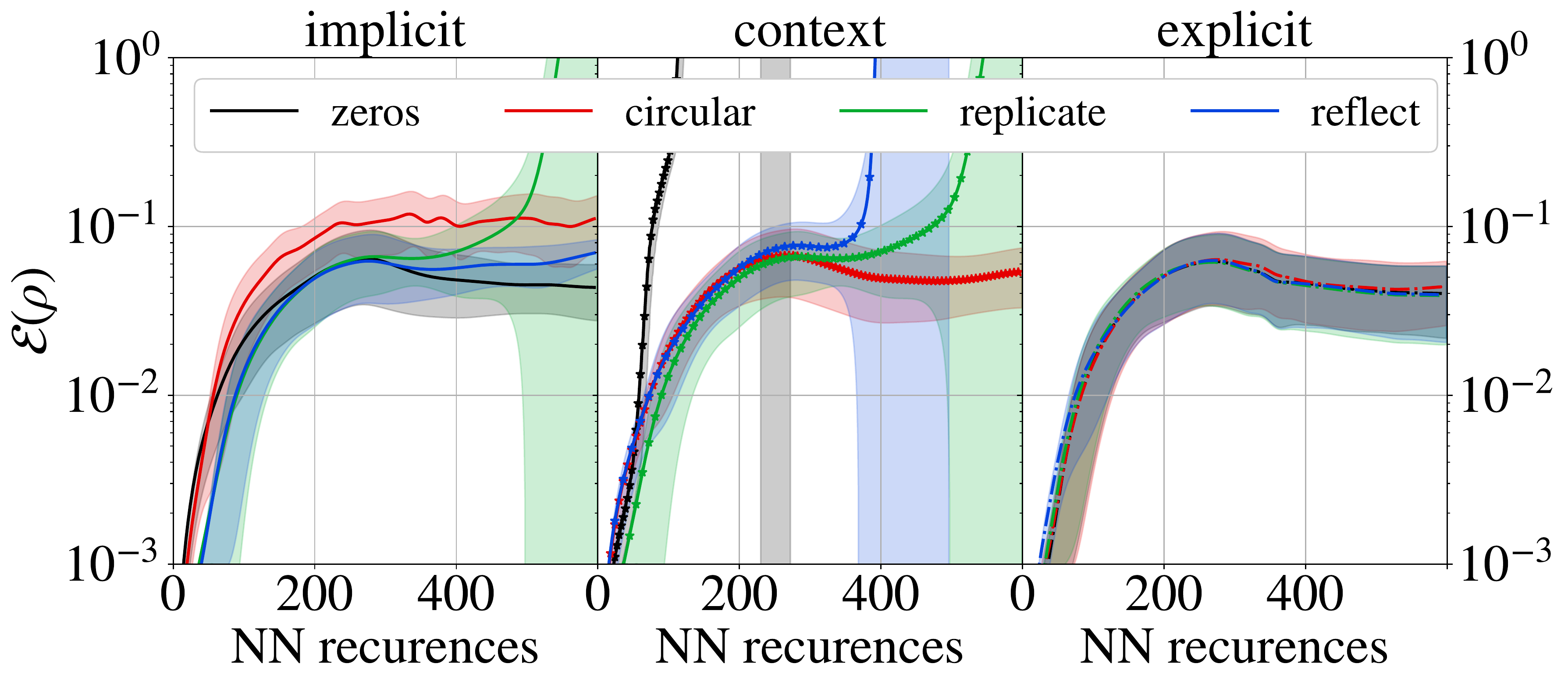}
    \caption{}
\label{fig:results_d3_error}
\end{subfigure}
\caption{
Relative errors for (a) D2 (periodic) and (b) D3 (absorbing) cases. Full lines represent the implicit strategy and lines and star markers the spatial context strategy.}
\label{fig:results_d2}
\end{figure}

The relative error evolution over time is depicted in Fig.\ref{fig:results_d2A}, for the implicit and spatial context methods. The explicit method is not employed here as it is equivalent to the implicit one: physical solvers employ additional ``ghost cells'' to wrap values from one boundary to another \cite{Mohan2020}. Also, reflect padding is not shown as it behaves very similarly to the replicate strategy. Results show that for both implicit and spatial context strategies, only circular padding is able to achieve acceptable error levels, with an average relative error of $15\%$ for the implicit case and $11\%$ at $it=600$ for the spatial context. The use of a padding strategy other than circular produces unphysical behavior at the boundaries, as the network is incapable of copying by itself the values arriving at one border to the opposite one. 


\begin{figure}[btp]
\begin{subfigure}[]{0.5\textwidth}
\settoheight{\tempdima}{\includegraphics[width=.2\linewidth]{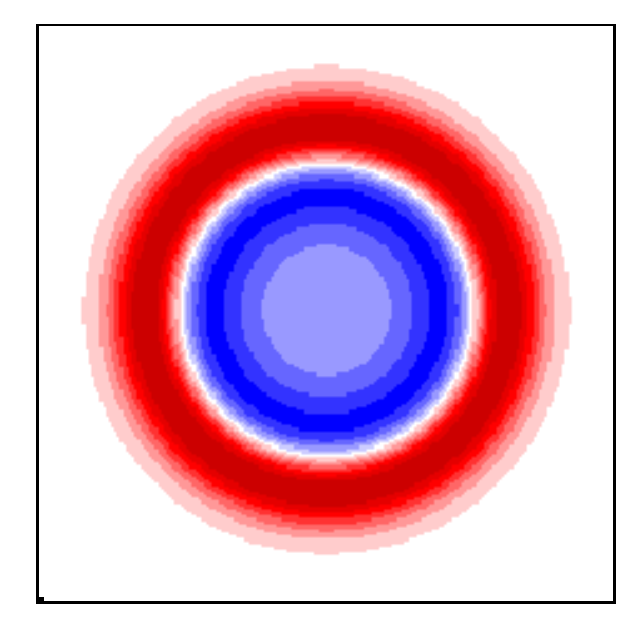}}%
\centering\begin{tabular}{@{}@{ }c@{}c@{}c@{ }c@{ }c@{}c@{}c@{}}
& \textbf{it=120} & \textbf{180} &\textbf{260} &\textbf{400}\\
\rowname{GT}&
\includegraphics[width=.2\linewidth]{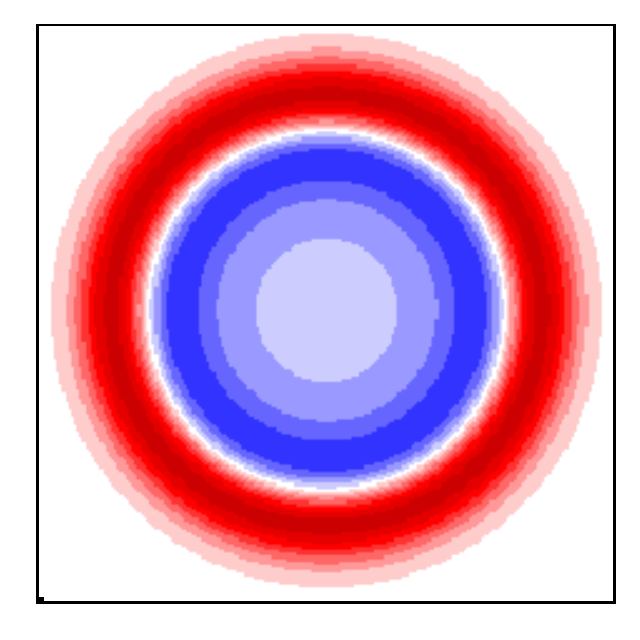}&
\includegraphics[width=.2\linewidth]{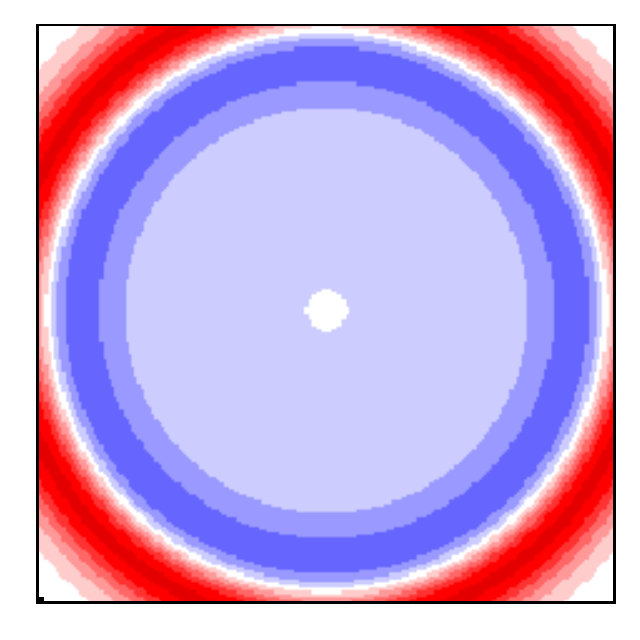}&
\includegraphics[width=.2\linewidth]{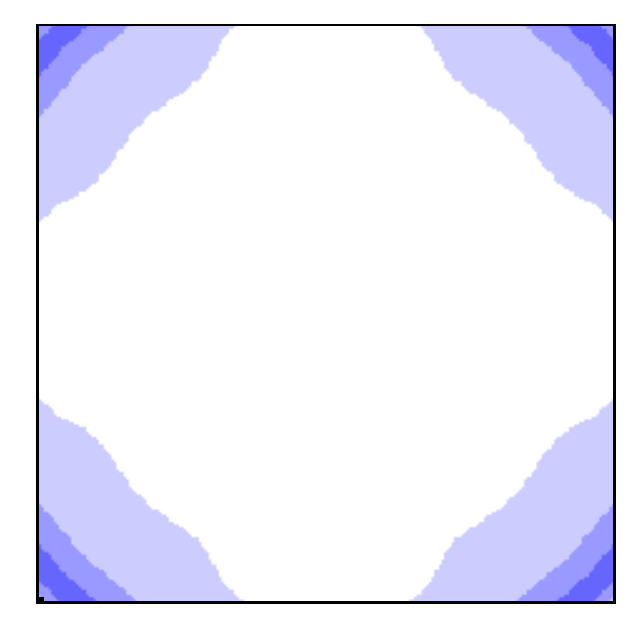}&
\includegraphics[width=.2\linewidth]{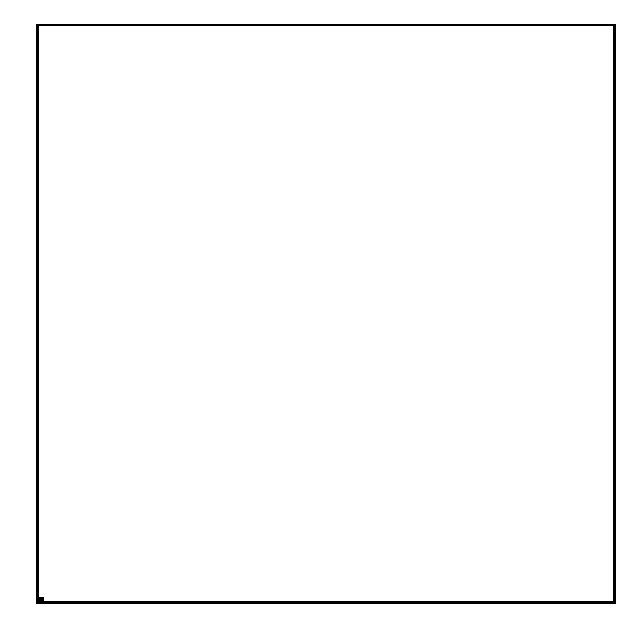}
\\[-1ex]
& & & & & \\
\rowname{Implicit}&
\includegraphics[width=.2\linewidth]{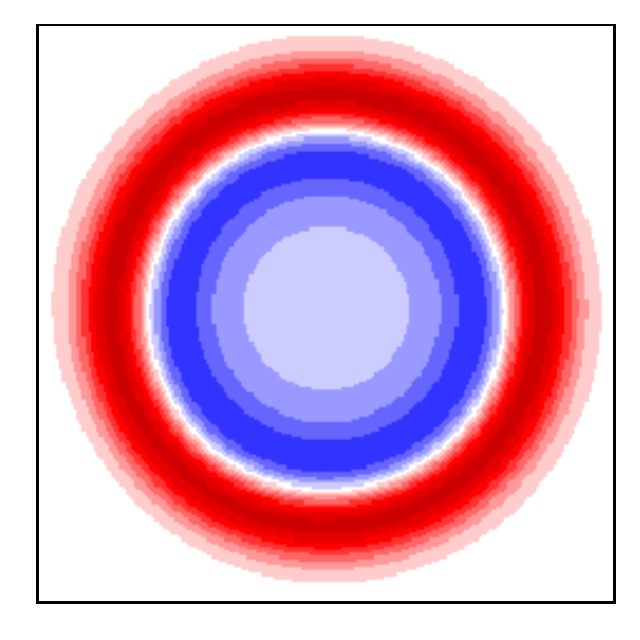}&
\includegraphics[width=.2\linewidth]{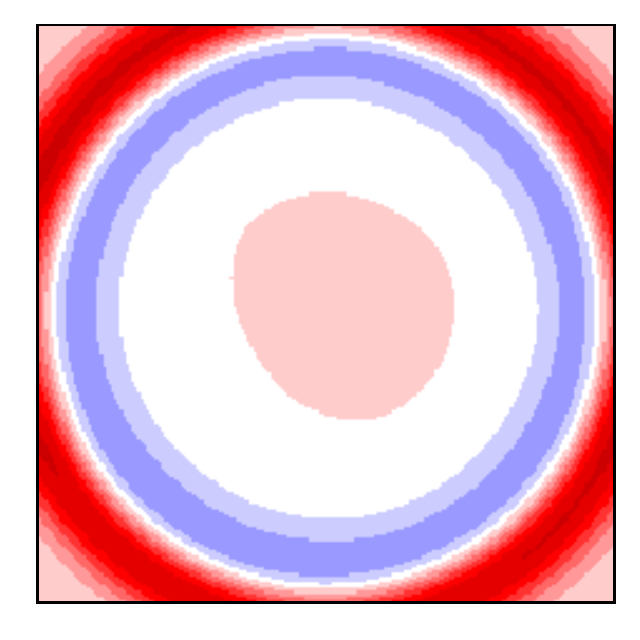}&
\includegraphics[width=.2\linewidth]{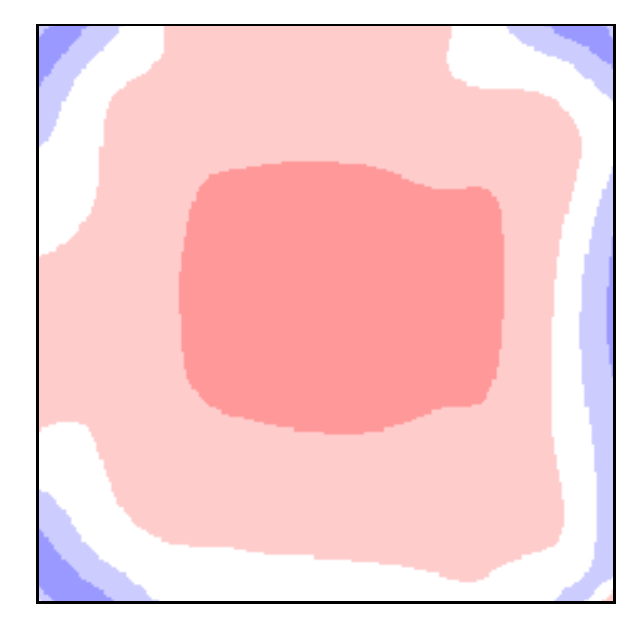}&
\includegraphics[width=.2\linewidth]{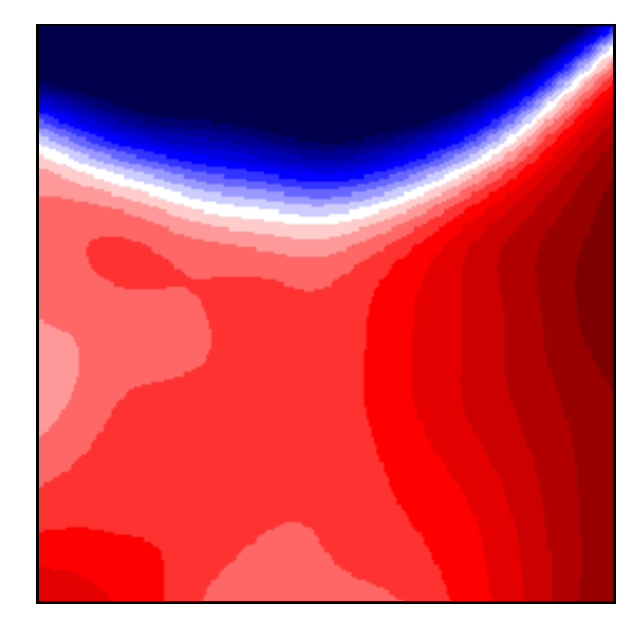}
\\[-1ex]
& & & & & \\
\end{tabular}
\end{subfigure}
\begin{subfigure}[]{0.5\textwidth}
\settoheight{\tempdima}{\includegraphics[width=.2\linewidth]{Figures/d3_abs/target/center/target_100.pdf}}%
\centering\begin{tabular}{@{}@{ }c@{}c@{}c@{ }c@{ }c@{}c@{}c@{}}
& \textbf{it=120} & \textbf{180} &\textbf{260} &\textbf{400}\\
\rowname{Context}&
\includegraphics[width=.2\linewidth]{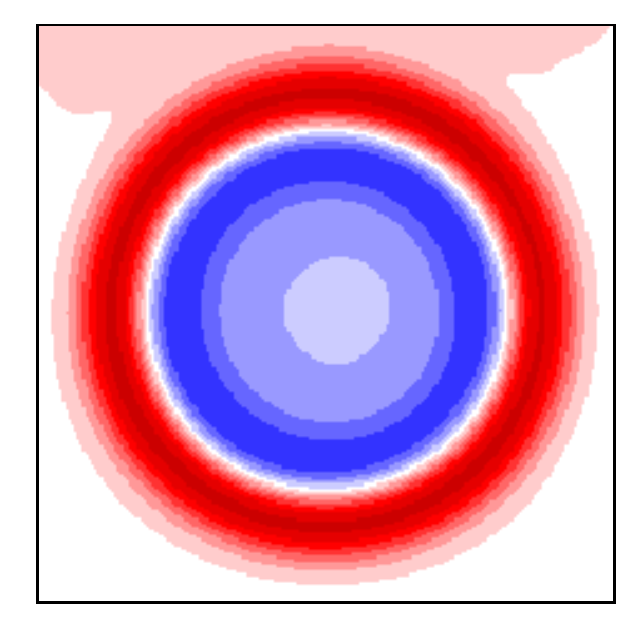}&
\includegraphics[width=.2\linewidth]{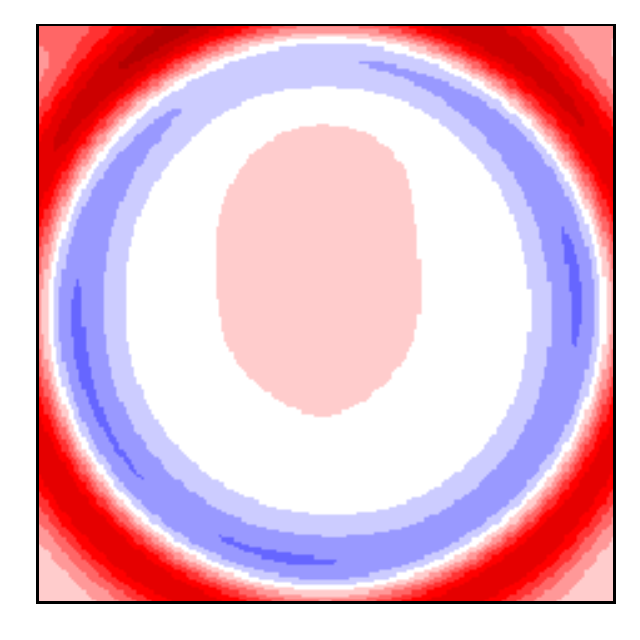}&
\includegraphics[width=.2\linewidth]{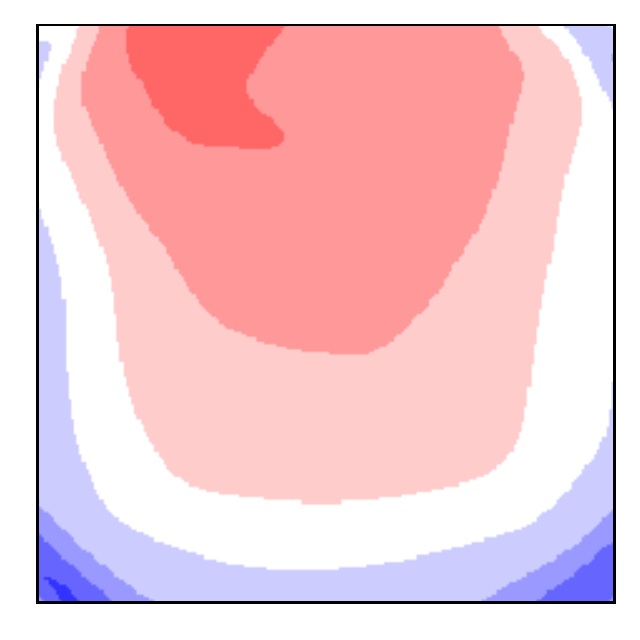}&
\includegraphics[width=.2\linewidth]{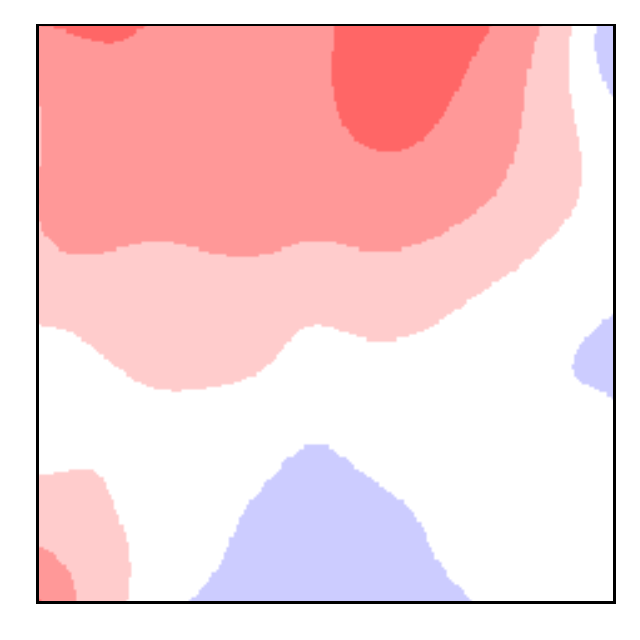}
\\[-1ex]
& & & & & \\
\rowname{Explicit}&
\includegraphics[width=.2\linewidth]{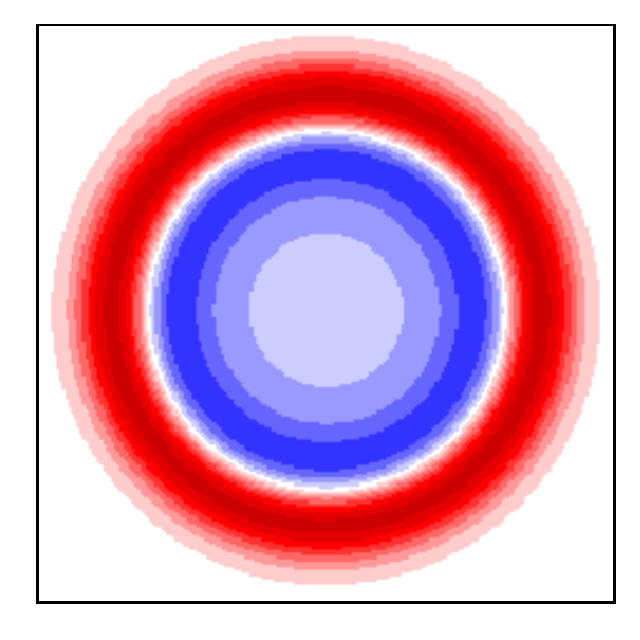}&
\includegraphics[width=.2\linewidth]{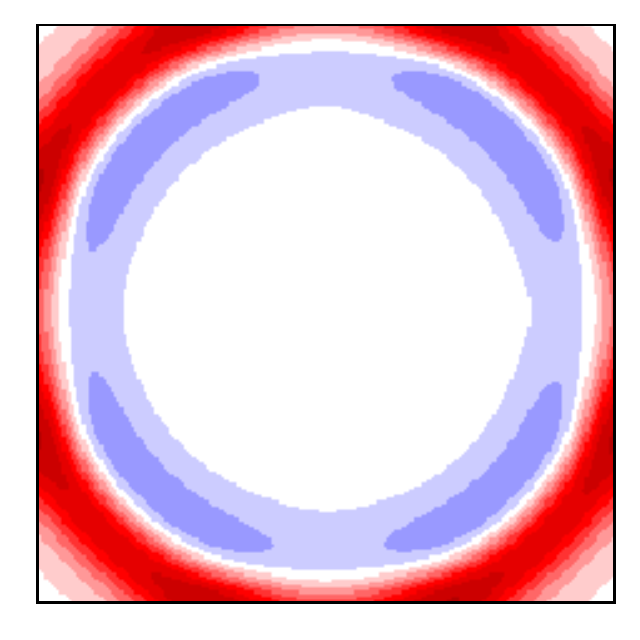}&
\includegraphics[width=.2\linewidth]{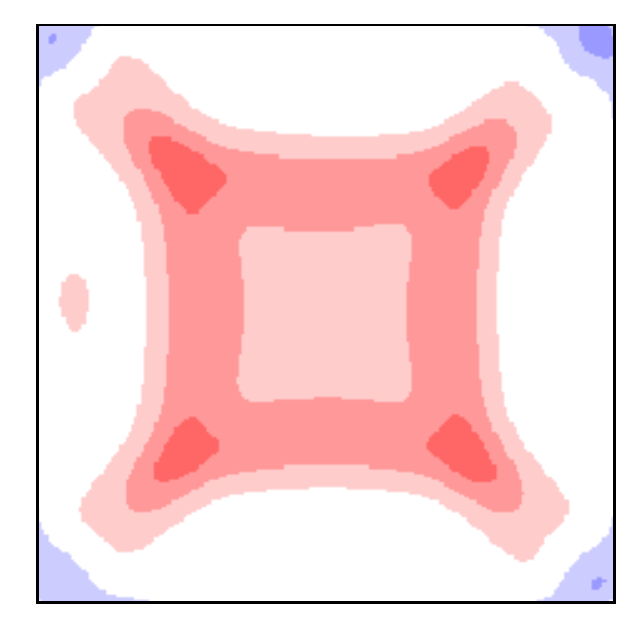}&
\includegraphics[width=.2\linewidth]{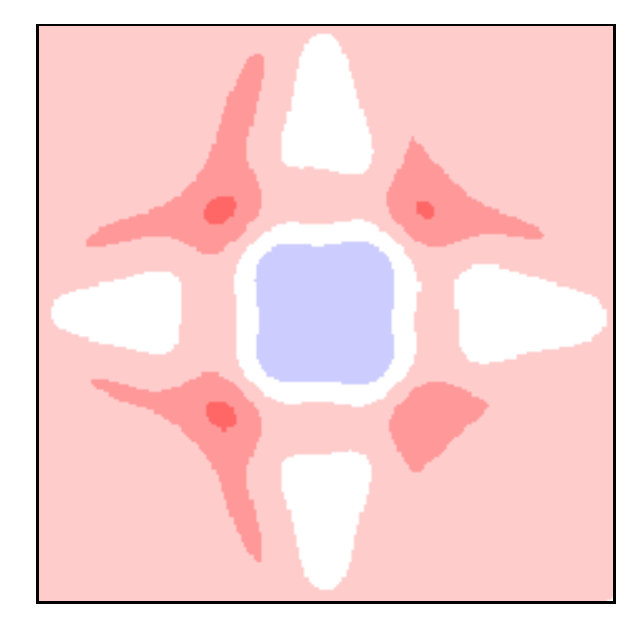}\\[-1ex]
& & & & & \\
\end{tabular}
\label{fig:results_d3A}
\end{subfigure}

\caption{Density fields for Dataset 3 and replicate padding, comparing the three investigated methods. The initial conditions is a centered pulse.}%
\label{fig:results_d3}
\end{figure}
\subsubsection{Wave Equation with Absorbing BCs}: The third test case corresponds to the neural network trained on dataset 3, with non-reflecting (absorbing) boundary conditions. This case is significantly more challenging than the two previous cases: the initial Gaussian pulse is expected to propagate into the far field and completely leave the computational domain. Thus, the underlying data distribution is changing over time: the initial acoustic energy $||\rho||_2$ tends towards $0$ as $t \to \infty$. The challenge for the network is to correctly propagate the initial pulses outside the domain without spurious reflections at boundaries.
As the signal energy tends towards zero, the error is now calculated relatively to the \textit{initial} density, i.e., $\mathcal{E}(\rho) = \sqrt{||\rho_{t} - \hat{\rho}_{t}||_2} /  |\rho_{t=0}|$. Similarly to previous experiments, 25 different initial conditions are employed for the auto-regressive tests.

\begin{figure}[tbp!]
\begin{subfigure}[]{0.5\textwidth}
\settoheight{\tempdima}{\includegraphics[width=.2\linewidth]{Figures/d3_abs/target/center/target_100.pdf}}%
\centering\begin{tabular}{@{}@{ }c@{}c@{}c@{ }c@{ }c@{}c@{}c@{}}
& \textbf{it=4} & \textbf{8} &\textbf{16} &\textbf{20}\\
\rowname{GT}&
\includegraphics[width=.2\linewidth]{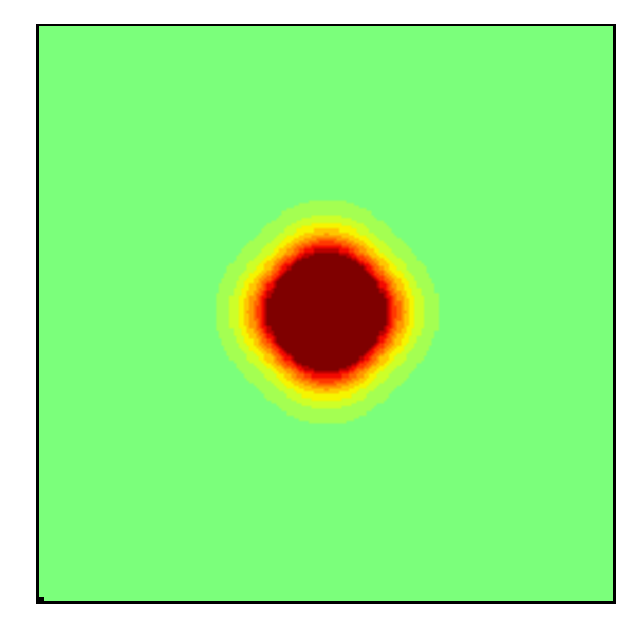}&
\includegraphics[width=.2\linewidth]{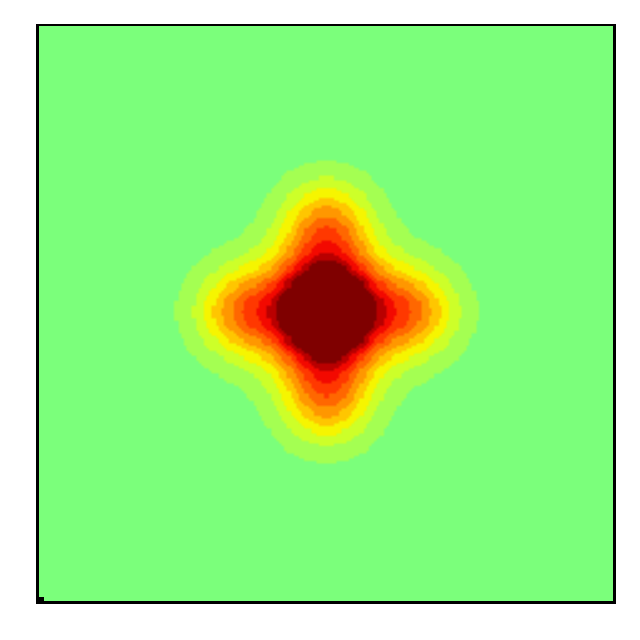}&
\includegraphics[width=.2\linewidth]{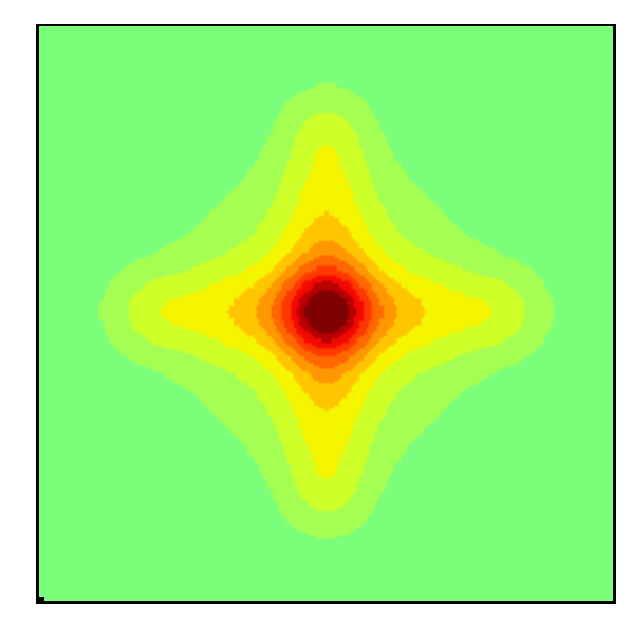}&
\includegraphics[width=.2\linewidth]{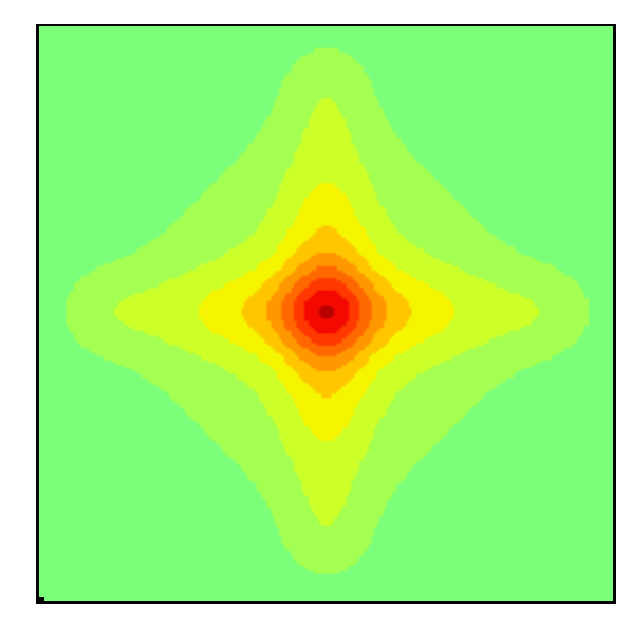}
\\[-1ex]
& & & & & \\
\rowname{Implicit}&
\includegraphics[width=.2\linewidth]{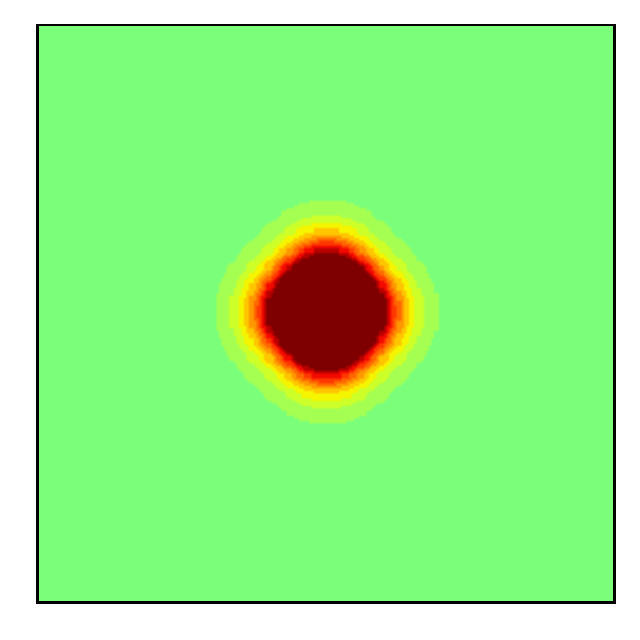}&
\includegraphics[width=.2\linewidth]{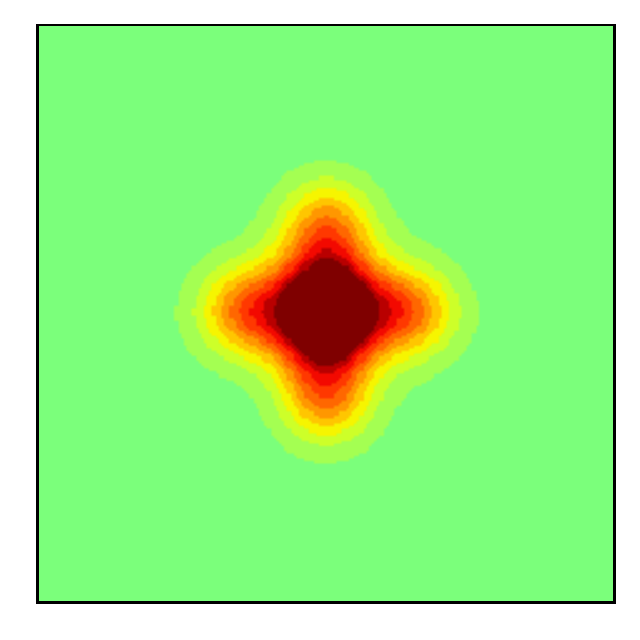}&
\includegraphics[width=.2\linewidth]{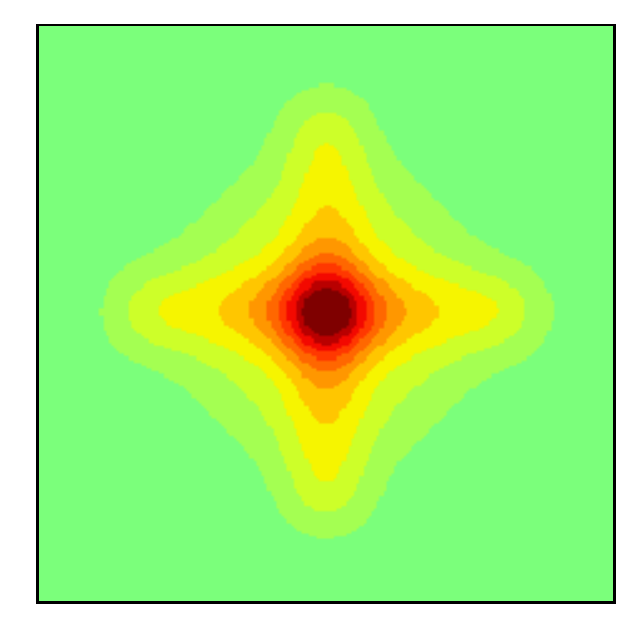}&
\includegraphics[width=.2\linewidth]{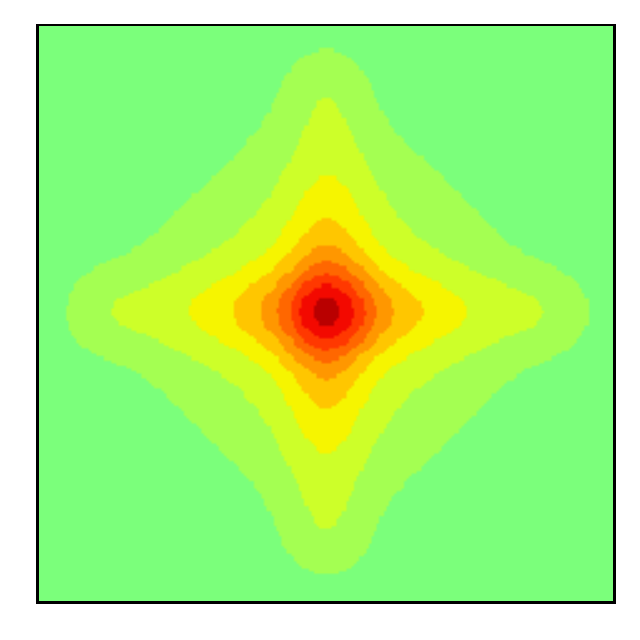}
\\[-1ex]
& & & & & \\
\end{tabular}
\end{subfigure}
\begin{subfigure}[]{0.5\textwidth}
\settoheight{\tempdima}{\includegraphics[width=.2\linewidth]{Figures/d3_abs/target/center/target_100.pdf}}%
\centering\begin{tabular}{@{}@{ }c@{}c@{}c@{ }c@{ }c@{}c@{}c@{}}
& \textbf{it=4} & \textbf{8} &\textbf{16} &\textbf{20}\\
\rowname{Context}&
\includegraphics[width=.2\linewidth]{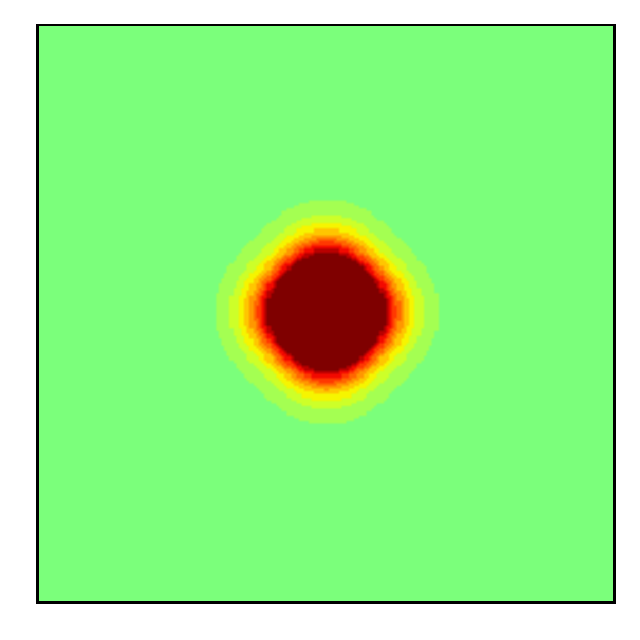}&
\includegraphics[width=.2\linewidth]{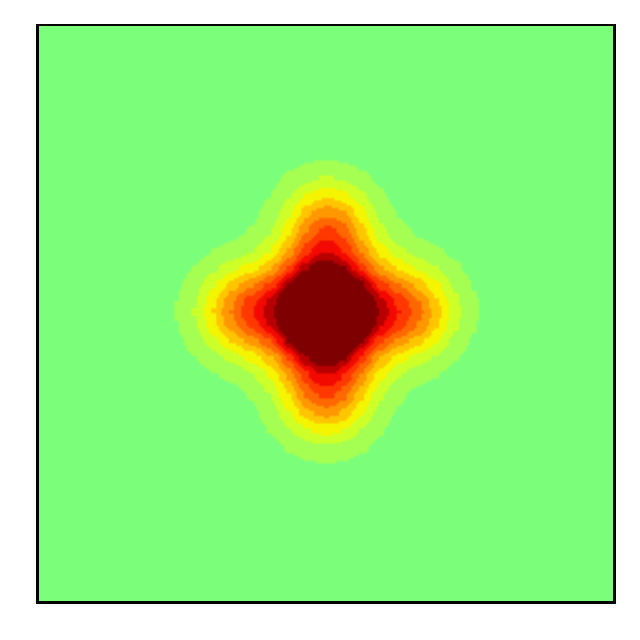}&
\includegraphics[width=.2\linewidth]{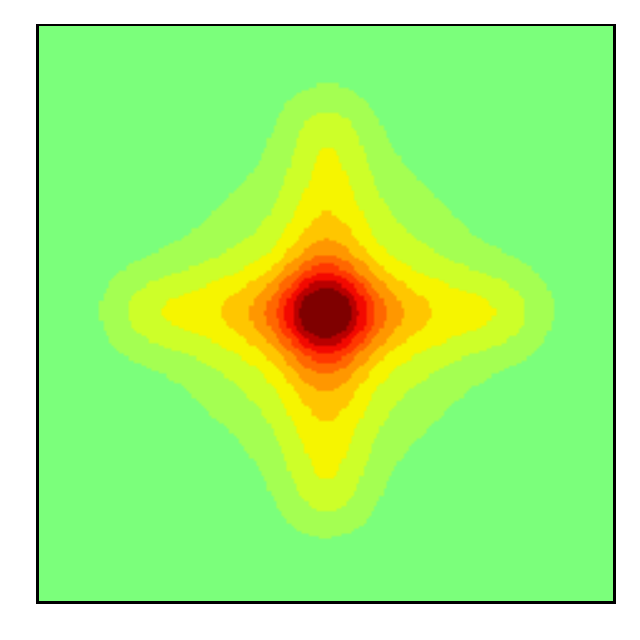}&
\includegraphics[width=.2\linewidth]{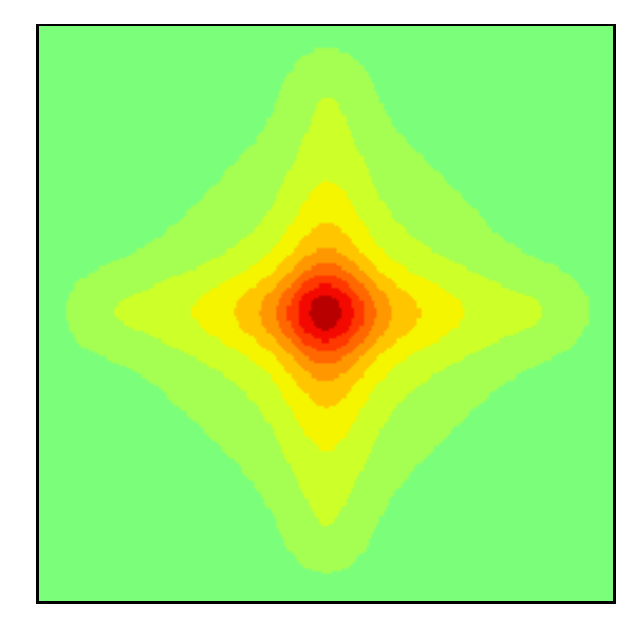}
\\[-1ex]
& & & & & \\
\rowname{Explicit}&
\includegraphics[width=.2\linewidth]{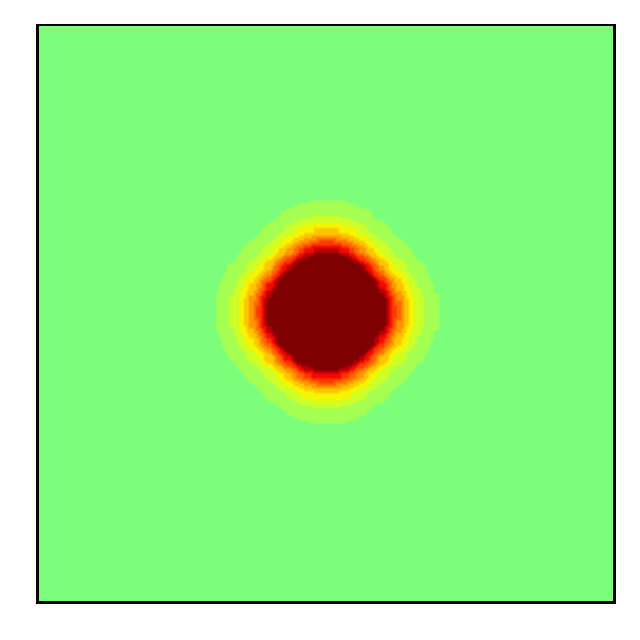}&
\includegraphics[width=.2\linewidth]{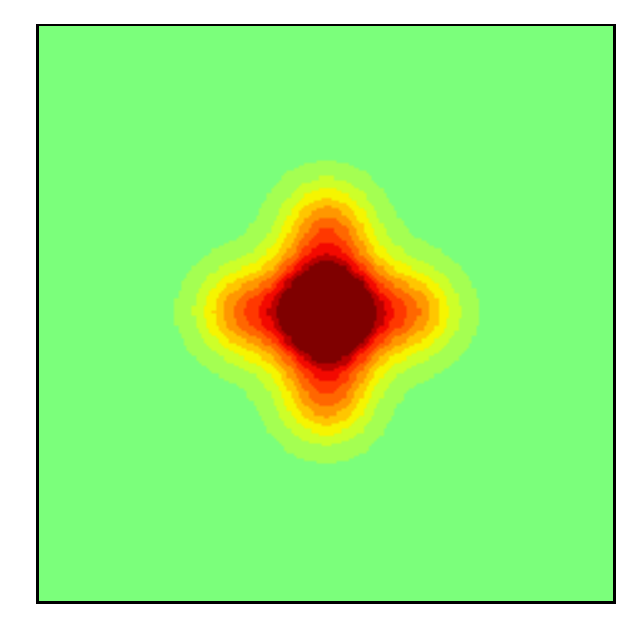}&
\includegraphics[width=.2\linewidth]{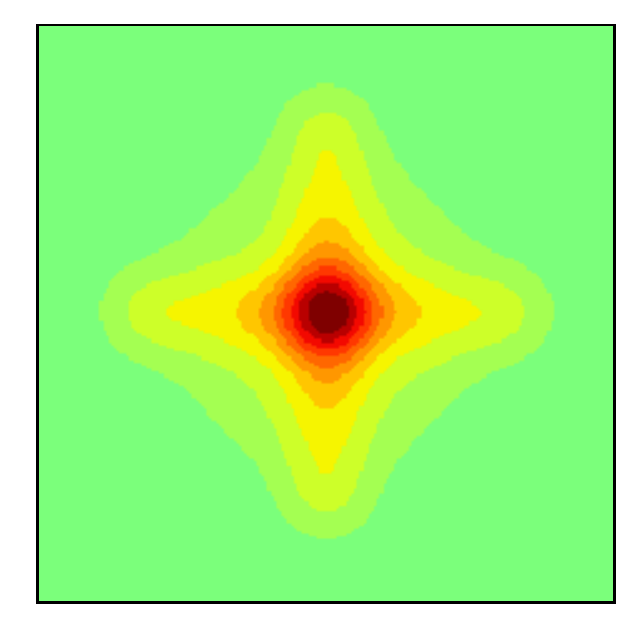}&
\includegraphics[width=.2\linewidth]{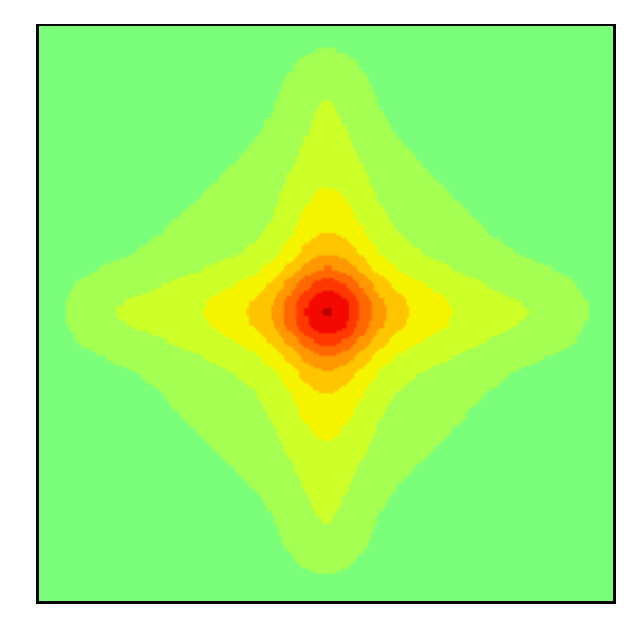}\\[-1ex]
& & & & & \\
\end{tabular}
\end{subfigure}
\caption{Temperature fields for Dataset 4 and replicate padding, comparing the three investigated methods. The initial condition is a centered pulse.}%
\label{fig:results_d4}
\end{figure}
%
%

For the explicit method, a local one-dimensional (LODI) non-reflecting equation is used to impose the values a the boundaries, which reads $ \partial_t \rho + c_0 \bm{\nabla} \rho \cdot \bm{n} = 0 $ \cite{Poinsot1992}. First order finite differences are employed to discretize both the spatial and temporal derivatives.

The evolution of the error for $600$ auto-regressive iterations is shown in Fig. \ref{fig:results_d3_error}. For both the implicit and spatial context cases, the results show a high variability between the different cases. While the implicit strategy with zero and reflect padding manage to produce low-error results, the other two padding strategies lead to diverging simulations. In contrast, when the spatial context is employed, the circular padding performs better, while the other three methods diverge. This unstable network behavior shows the complexity of the non-reflecting case in comparison to the previous ones. Instabilities can be directly related to the appearance of artifacts when the pulse impinges the walls, as can be seen in Fig. \ref{fig:results_d3}: before the pulse arrival at the wall ($it=120$), all methods show stable and accurate predictions, while larger errors are shown after the first interaction with the BCs ($it > 160)$. Such artifacts lead in some cases to the unbounded growth of the density amplitude, leading to instabilities.

Interestingly, the explicit encoding of boundaries has an important stabilizing effect: all four padding methods show a very similar error evolution. Figures \ref{fig:results_d3} show the main differences between implicit, spatial context and explicit approaches for the same padding (replicate): the first two cases initially damp the outgoing waves more efficiently. However, some unphysical reflections (shown by the asymmetry of density the fields) lead to artifacts remaining in the computational domain, eventually leading to instabilities in the implicit case. In the explicit case, even though reflections also exist, they follow a symmetric pattern, which corresponds very closely to the one found for reflecting walls, as seen in Fig. \ref{fig:results_d1}. This suggests that the explicit encoding adds physical consistency to the network BC treatment. The error in that approach arises mainly from the lack of proper BC modeling, as the aforementioned LODI hypothesis cannot handle efficiently two-dimensional effects, typical of Gaussian pulses, for example at corners. Future research should reach out towards the treatment of such transverse effects \cite{Poinsot1992}. This demonstrates that blending prior physics knowledge when handling boundary conditions can improve CNNs accuracy and robustness.

\begin{figure}[t]
    \centering
    \includegraphics[width=\textwidth]{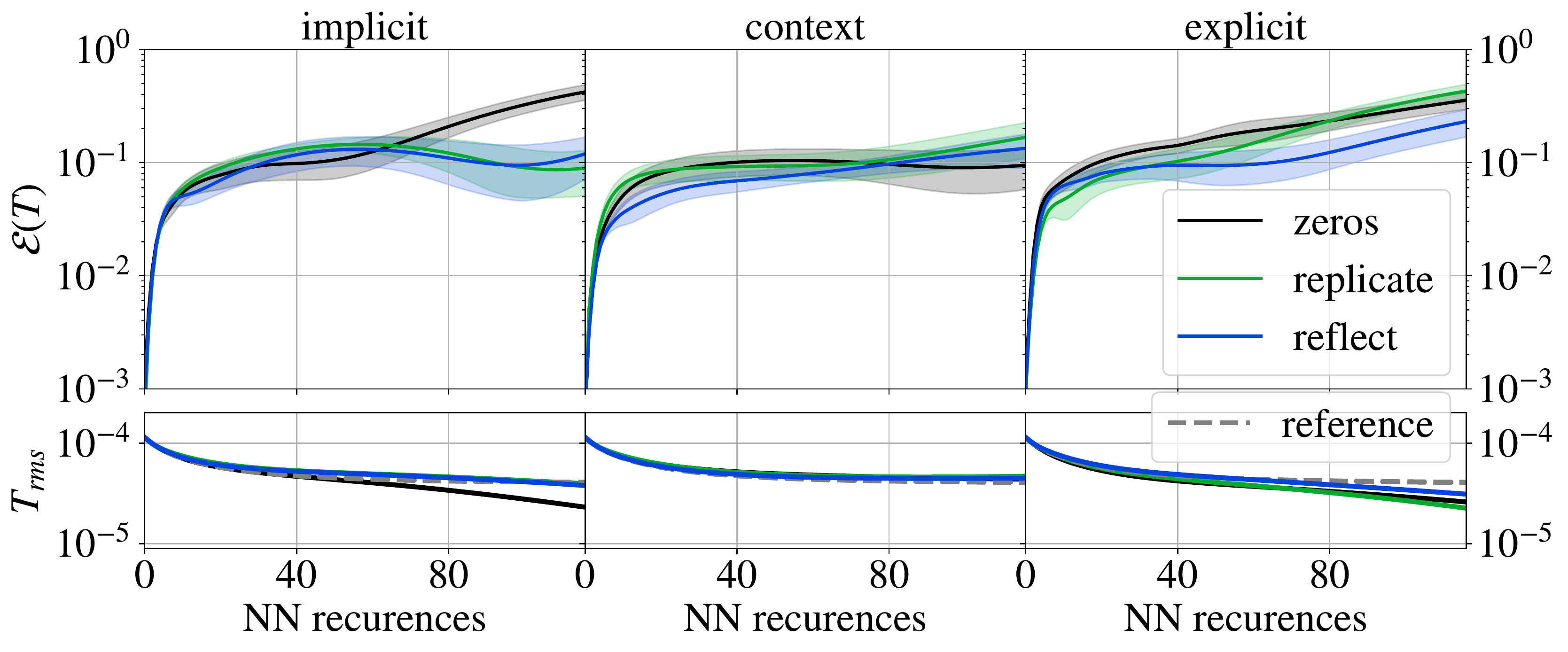}
\caption{Results for case D4. (Top) Relative MSE on temperature fields, (bottom) temporal evolution of the spatially averaged temperature $T_{rms}$. }
\label{fig:error_d4}
\end{figure}

\subsubsection{Heat Equation with Adiabatic Boundary}: The last studied application corresponds to the diffusion of temperature pulses, modeled by the heat equation (\ref{eq:heatEq}), when employing Neumann boundary conditions in the temperature (adiabatic walls). The same testing strategy is employed, by simulating 25 initial conditions for $T=120$ neural network recurrences. Two metrics are employed to analyze the network, the relative mean square error $\mathcal{E}(T) = \sqrt{||T_{t} - \hat{T}_{t}||_2} /  \sqrt{||T_{t}||_2} $ and the evolution of the average spatial temperature $T_{rms}$ 
that tends to a constant value in time, as no heat flux exists at the adiabatic walls. 

Results in Figs.~\ref{fig:results_d4} and \ref{fig:error_d4} confirm some of the previously made observations for Neumann BCs (reflecting acoustic case): the implicit padding strategy seems appropriate to correctly reproduce the dynamics, when those remain simple, such as in the diffusion of temperature fields. However, the error evolution shows that the use of padding that mimics the physical boundary condition (e.g. replicate) results in a lower long-term error, as well as a closer approximation to the $T_{rms}$ constant level. Furthermore, the 
use of additional spatial context reduces the variability of the different padding choices, thus confirming the interest of employing such an additional input to make the predictions more robust to boundary condition effects. Here the explicit strategy does not results in an improved accuracy with respect to the baseline methods. The use of such an additional a-priori encoding of boundary conditions may only be justified in the presence of complex conditions, such as the one presented previously for the non-reflecting acoustics case.
\section{Conclusion}

This paper presents an exhaustive comparison between several available methods to treat boundary conditions in fully convolutional neural networks for spatio-temporal regression, in the context of hyperbolic and parabolic PDEs. Such temporal regression tasks are highly sensitive to the well-posedness of boundary conditions, as small localized errors can propagate in time and space, producing instabilities in some cases. The characterization of such boundaries is crucial to improve the neural network accuracy. 

The main outcomes are summarized next: employing padding alone yields accurate results only when the chosen padding is compatible with the underlying data. The addition of a spatial context channel seems to increase the robustness of the network in simple cases (Neumann boundaries), but fails for the more complex non-reflecting boundary case. Finally, the explicit encoding of boundaries, which enforces some physics constraints on border pixels, clearly demonstrates its superiority in such cases, allowing to design more robust neural networks. Such an approach should be further investigated in order to understand its coupling with the neural network behavior, and its extension to problems with several types of boundary conditions.

\bibliographystyle{splncs04}

\begin{thebibliography}{10}
\providecommand{\url}[1]{\texttt{#1}}
\providecommand{\urlprefix}{URL }
\providecommand{\doi}[1]{https://doi.org/#1}

\bibitem{Alguacil2020b}
Alguacil, A., Bauerheim, M., Jacob, M.C., Moreau, S.: {Predicting the
  Propagation of Acoustic Waves using Deep Convolutional Neural Networks}. In:
  AIAA Aviation Forum. p.~2513. Reston, VA (2020)

\bibitem{Alsallakh2021}
Alsallakh, B., Kokhlikyan, N., Miglani, V., Yuan, J., Reblitz-Richardson, O.:
  {Mind the Pad - CNNs can develop blind spots}. In: 9th International
  Conference on Learning Representations (ICLR). Vienna, Austria (2021)

\bibitem{Fotiadis2020}
Fotiadis, S., Pignatelli, E., Bharath, A.A., {Lino Valencia}, M., Cantwell,
  C.D., Storkey, A.: {Comparing Recurrent and Convolutional Neural Networks for
  Predicting Wave Propagation}. ICLR 2020 Workshop on Deep Learning and
  Differential Equations  (2020)

\bibitem{Gao2021}
Gao, H., Sun, L., Wang, J.X.: {PhyGeoNet: Physics-Informed Geometry-Adaptive
  Convolutional Neural Networks for Solving Parametric PDEs on Irregular
  Domain}. Journal of Computational Physics  \textbf{428},  110079 (mar 2021)

\bibitem{Geneva2019}
Geneva, N., Zabaras, N.: {Modeling the Dynamics of PDE Systems with
  Physics-Constrained Deep Auto-Regressive Networks}. Journal of Computational
  Physics  \textbf{403} (2019)

\bibitem{Hamey2015}
Hamey, L.G.: {A Functional Approach to Border Handling in Image Processing}.
  In: International Conference on Digital Image Computing: Techniques and
  Applications, (DICTA). Adelaide, Australia (2015)

\bibitem{Innamorati2020}
Innamorati, C., Ritschel, T., Weyrich, T., Mitra, N.J.: {Learning on the Edge:
  Investigating Boundary Filters in CNNs}. International Journal of Computer
  Vision  \textbf{128}(4),  773--782 (apr 2020)

\bibitem{Islam2020}
Islam, A., Jia, S., Bruce, N.D.B.: {How much Position Information do
  Convolutional Neural Networks encode?} In: 8th International Conference on
  Learning Representations (ICLR). Addis Ababa, Ethiopia (2020)

\bibitem{Kayhan}
Kayhan, O.S., {Van Gemert}, J.C.: {On Translation Invariance in CNNs:
  Convolutional Layers can Exploit Absolute Spatial Location}. In: Proceedings
  of the IEEE/CVF Conference on Computer Vision and Pattern Recognition (CVPR).
  pp. 14274--14285. Virtual Event (2020)

\bibitem{Latt2020}
Latt, J., Malaspinas, O., Kontaxakis, D., Parmigiani, A., Lagrava, D., Brogi,
  F., Belgacem, M.B., Thorimbert, Y., Leclaire, S., Li, S., Marson, F., Lemus,
  J., Kotsalos, C., Conradin, R., Coreixas, C., Petkantchin, R., Raynaud, F.,
  Beny, J., Chopard, B.: {Palabos: Parallel Lattice Boltzmann Solver}.
  Computers and Mathematics with Applications  (apr 2020)

\bibitem{Lee2019a}
Lee, S., You, D.: {Data-driven prediction of unsteady flow over a circular
  cylinder using deep learning}. Journal of Fluid Mechanics  \textbf{879}(1),
  217--254 (nov 2019)

\bibitem{Liu2018a}
Liu, G., Shih, K.J., Wang, T.C., Reda, F.A., Sapra, K., Yu, Z., Tao, A.,
  Catanzaro, B.: {Partial Convolution based Padding}. arXiv preprint
  arXiv:1811.11718  (nov 2018)

\bibitem{Liu2008}
Liu, R., Jia, J.: {Reducing boundary artifacts in image deconvolution}. In:
  15th IEEE International Conference on Image Processing. pp. 505--508. San
  Diego, CA (2008)

\bibitem{Liu2018}
Liu, R., Lehman, J., Molino, P., Such, F.P., Frank, E., Sergeev, A., Yosinski,
  J.: {An intriguing failing of convolutional neural networks and the CoordConv
  solution}. In: Advances in Neural Information Processing Systems 31. pp.
  9605--9616. Montr{\'{e}}al, Canada (2018)

\bibitem{Mathieu2016a}
Mathieu, M., Couprie, C., LeCun, Y.: {Deep multi-scale video prediction beyond
  mean square error}. 4th International Conference on Learning Representtions,
  ICLR  (nov 2016)

\bibitem{Mohan2020}
Mohan, A.T., Lubbers, N., Livescu, D., Chertkov, M.: {Embedding Hard Physical
  Constraints in Neural Network Coarse-Graining of 3D Turbulence}. In: ICLR
  2020 Workshop Tackling Climate Change with Machine Learning. arXiv (jan 2020)

\bibitem{Poinsot1992}
Poinsot, T.J., Lele, S.K.: {Boundary conditions for direct simulations of
  compressible viscous flows}. Journal of Computational Physics
  \textbf{101}(1),  104--129 (1992)

\bibitem{Raissi2019a}
Raissi, M., Perdikaris, P., Karniadakis, G.E.: {Physics-informed neural
  networks: A deep learning framework for solving forward and inverse problems
  involving nonlinear partial differential equations}. Journal of Computational
  Physics  \textbf{378},  686--707 (2019)

\bibitem{Schubert2019}
Schubert, S., Neubert, P., Poschmann, J., Pretzel, P.: {Circular convolutional
  neural networks for panoramic images and laser data}. In: 2019 IEEE
  Intelligent Vehicles Symposium (IV). pp. 653--660. Paris, France (2019)

\end{thebibliography}

\end{document}